\documentclass{article}

\usepackage{microtype}
\usepackage{graphicx}
\usepackage{subfigure}
\usepackage{booktabs} 
\usepackage{enumitem}
\usepackage[sort,compress]{natbib}
\usepackage{xcolor}
\usepackage[hyphens,spaces,obeyspaces]{url}

\usepackage{hyperref}

\usepackage[accepted]{icml2024}

\usepackage{amsmath}
\usepackage{amssymb}
\usepackage{mathtools}
\usepackage{amsthm}
\usepackage{xspace}

\usepackage[capitalize,noabbrev]{cleveref}

\usepackage{macros}
\usepackage{makecell}

\definecolor{green}{HTML}{009E73}

\theoremstyle{plain}

\theoremstyle{definition}

\theoremstyle{remark}

\usepackage[textsize=tiny]{todonotes}

\icmltitlerunning{Improving Token-Based World Models with Parallel Observation Prediction}

\begin{document}

\twocolumn[
\icmltitle{Improving Token-Based World Models with Parallel Observation Prediction}




\begin{icmlauthorlist}
\icmlauthor{Lior Cohen}{technion}
\icmlauthor{Kaixin Wang}{technion}
\icmlauthor{Bingyi Kang}{bd}
\icmlauthor{Shie Mannor}{technion}
\end{icmlauthorlist}

\icmlaffiliation{technion}{Technion – Israel Institute of Technology}
\icmlaffiliation{bd}{ByteDance}

\icmlcorrespondingauthor{Lior Cohen}{liorcohen5@campus.technion.ac.il}

\icmlkeywords{Machine Learning, ICML, World Model, Reinforcement Learning}

\vskip 0.3in
]



\printAffiliationsAndNotice{}  

\begin{abstract}
Motivated by the success of Transformers when applied to sequences of discrete symbols, token-based world models (TBWMs) were recently proposed as sample-efficient methods.
In TBWMs, the world model consumes agent experience as a language-like sequence of tokens, where each observation constitutes a sub-sequence.
However, during imagination, the sequential token-by-token generation of next observations results in a severe bottleneck, leading to long training times, poor GPU utilization, and limited representations.
To resolve this bottleneck, we devise a novel Parallel Observation Prediction (POP) mechanism.
POP augments a Retentive Network (RetNet) with a novel forward mode tailored to our reinforcement learning setting.
We incorporate POP in a novel TBWM agent named REM (Retentive Environment Model), showcasing a 15.4x faster imagination compared to prior TBWMs.
REM attains superhuman performance on 12 out of 26 games of the Atari 100K benchmark, while training in less than 12 hours. 
Our code is available at \url{https://github.com/leor-c/REM}.

\end{abstract}

\section{Introduction}




Sample efficiency remains a central challenge in reinforcement learning (RL) due to the substantial data demands of successful RL algorithms~\citep{mnih2015human, silver2016alphaGo, Schrittwieser2020, openai2019dota2, Vinyals2019alphaStar}.
One prominent model-based approach for addressing this challenge is known as world models. 
In world models, the agent’s learning is solely based on simulated interaction data produced by a learned model of the environment through a process called imagination.
World models are gaining increasing popularity due to their effectiveness, particularly in visual environments~\cite{hafner2023mastering}.

\begin{figure}[t]
    \centering
    \includegraphics[width=\linewidth]{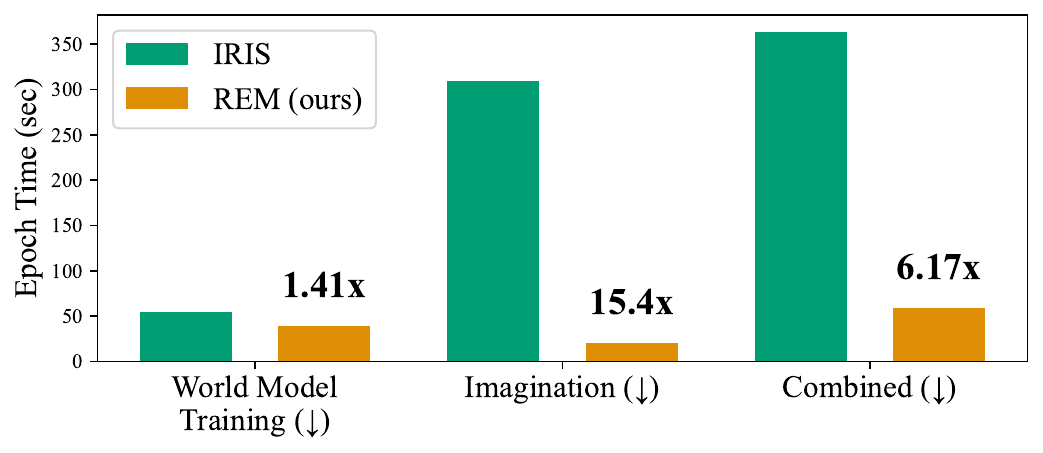}
    \includegraphics[width=\linewidth]{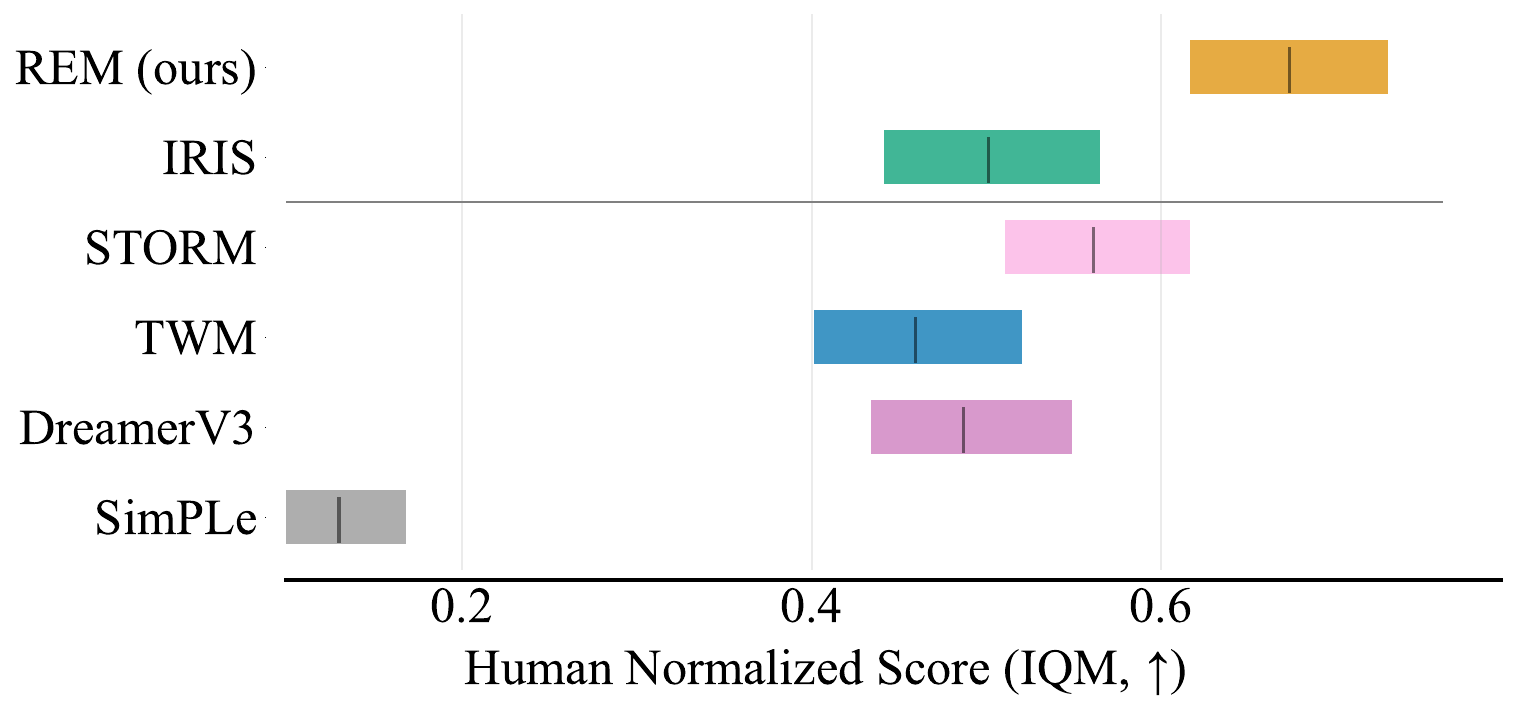}
    \vspace{-20pt}
    \caption{\textbf{Top:} comparison between the run times of token-based world model agents (IRIS and REM) during the world model training and imagination (actor-critic training). 
    \textbf{Bottom:} interquantile mean (IQM) human-normalized score comparison between REM and state-of-the-art baselines on the Atari 100K benchmark with 95\% stratified bootstrap confidence intervals \cite{Agarwal2021rliable}. A line separates token-based methods from other baselines.}
    \label{fig:first-page-results}
\end{figure}

%

\begin{figure*}[t]
    \centering
    \includegraphics[width=0.9\linewidth]{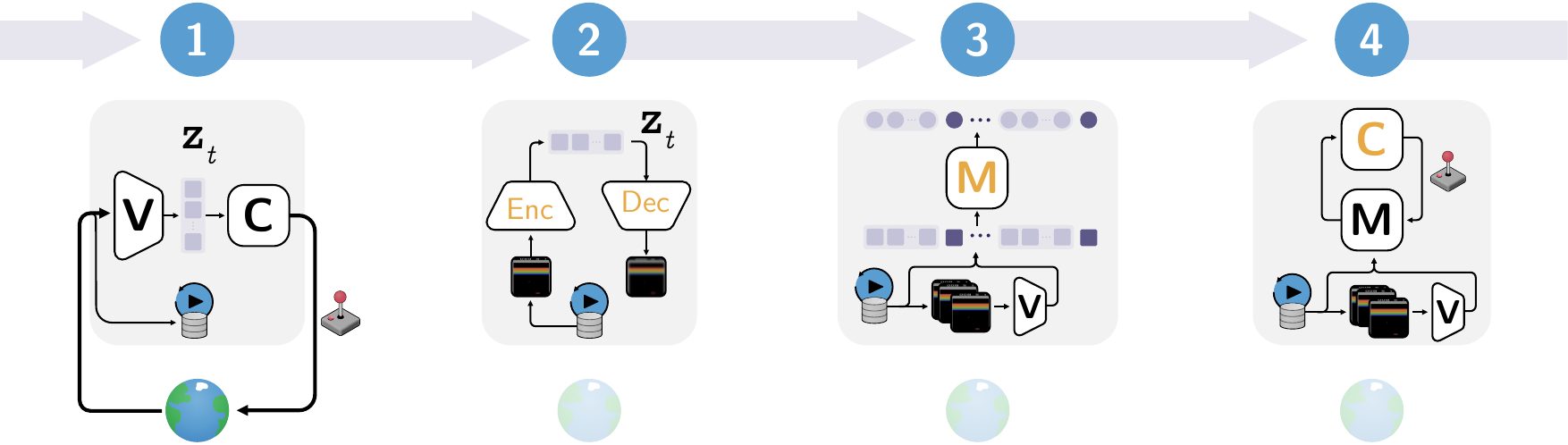}
    \caption{An overview of REM's training cycle. Each epoch has 4 steps: experience collection (1), tokenizer training (2), world model training (3), and controller training in imagination (4). Orange color represents component(s) that undergo training. Blue squares denote token inputs, where light blue is used for observation tokens and dark blue for actions. Replay buffer data at steps 3 and 4 contains observations, actions, rewards, and termination signals. }
    \label{fig:overview}
\end{figure*}

In recent years, attention-based sequence models, most notably the Transformer architecture \cite{Vaswani2017Attention}, demonstrated unmatched performance in language modeling tasks \cite{Devlin2019BERT, brown2020LMsFewShotLearners, openai2023sparksOfAGI, Touvron2023Llama2O}.
The notable success of these models when applied to sequences of discrete tokens sparked motivation to employ these architectures to other data modalities by learning appropriate token-based representations.
In computer vision, discrete representations are becoming a mainstream approach for various tasks \cite{vanDenOord2017vqvae, dosovitskiy2021an, esser2021tamingVQGAN, li2023mage}.
In RL, \emph{token-based} world models were recently explored in visual environments \cite{micheli2022transformers}.
The visual perception module in these methods is called a tokenizer, as it maps image observations to sequences of discrete symbols.
This way, agent interaction is translated into a language-like sequence of discrete tokens, which are processed individually by the world model.

During imagination, to generate the tokens of the next observation with the auto-regressive model, the prediction is carried sequentially token-by-token.
Effectively, this highly-sequential computation results in a severe bottleneck that pronouncedly hinders token-based approaches. 
Consequently, this bottleneck practically caps the length of the observation token sequences which in turn degrades performance.
This limitation renders current token-based methods impractical for complex problems.

In this paper, we present Parallel Observation Prediction (POP), a novel mechanism that resolves the imagination bottleneck of token based world models  (TBWMs).
With POP, the enire next observation token sequence is generated in parallel during world model imagination.
At its core, POP augments a Retentive Network (RetNet) sequence model \cite{sun2023retentive} with a novel forward mode devised for retaining world model training efficiency.
Additionally, we present REM (Retentive Environment Model), a TBWM agent driven by a POP-augmented RetNet architecture.

Our main contributions are summarized as follows:
\begin{itemize}
    \item We propose Parallel Observation Prediction (POP), a novel mechanism that resolves the inference bottleneck of current token-based world models while retaining performance.
    \item We introduce REM, the first world model approach that incorporates the RetNet architecture. Our experiments provide first evidence of RetNet's performance in an RL setting.
    \item We evaluate REM on the Atari 100K benchmark, demonstrating the effectiveness of POP. POP leads to a 15.4x speed-up at imagination and trains in under 12 hours, while outperforming prior TBWMs. 
\end{itemize}

%





\section{Method}

\textbf{Notations.}
We consider the Partially Observable Markov Decision Process (POMDP) setting 
with  image observations $\obs_t \in \obsSet \subseteq \mathbb{R}^{h \times w \times 3}$,
discrete actions $\action_t \in \actionsSet$, 
scalar rewards $\reward_t \in \mathbb{R}$,
episode termination signals $\doneSgnl_{t} \in \{0,1\}$,
dynamics $\obs_{t+1}, \reward_{t}, \doneSgnl_{t} \sim p(\obs_{t+1}, \reward_{t}, \doneSgnl_{t} \vert \obs_{\leq t}, \action_{\leq t})$,
and discount factor $\discountF$.
The objective is to learn a policy $\policy$ such that for every situation the output $\policy(\action_t | \obs_{\leq t}, \action_{< t})$ is optimal w.r.t. the expected discounted sum of rewards from that situation $\E [ \sum_{\tau=0}^{\infty} \discountF^{\tau} \rewardRv_{t+\tau} ]$ under the policy $\policy$. 

\subsection{Overview}
\label{subsec:method-wm}

REM builds on IRIS \cite{micheli2022transformers}, and similar to most prior works on world models for pixel input~\citep{hafner2021dreamerv2,Kaiser2020Model,hafner2023mastering}, REM follows a $\Tokenizer$-$\WM$-$\Controller$ structure~\citep{ha2018worldmodels}:
a $\Tokenizer$isual perception module that compresses observations into compact latent representations, a predictive $\WM$odel that captures the environment's dynamics, and a $\Controller$ontroller that learns to act to maximize return.
Additionally, a replay buffer is used to store environment interaction data.
An overview of REM's training cycle is presented in Figure \ref{fig:overview}.
A pseudo-code algorithm of REM is presented in Appendix \ref{sec:rem-algorithm-pseudo-code}. 



\paragraph{$\Tokenizer$ - Tokenizer}

We instantiate the visual perception component as a tokenizer, mapping input observations into latent tokens.
Following \cite{micheli2022transformers}, the tokenizer is a VQ-VAE discrete auto-encoder~\cite{vanDenOord2017vqvae, esser2021tamingVQGAN} comprised of an encoder, a decoder, and an embedding table.
The embedding table $\tknzrCodebook=\{\tknEmb_i\}_{i=1}^N\in\mathbb{R}^{N\times d}$ consists of $N$ trainable vectors.
The encoder first maps an input image $\obs_t$ to a sequence of $d$-dimensional latent vectors $(\tknzrLatent^1_t,\tknzrLatent^2_t,\cdots,\tknzrLatent^K_t)$.
Then, each latent vector $\tknzrLatent^k_t \in \mathbb{R}^d$ is mapped to the index of the nearest embedding in $\tknzrCodebook$, , $\token_{t}^{k} = \argmin_i \Vert \tknzrLatent_{t}^{k} - \tknEmb_{i} \Vert$.
Such indices are called \textit{tokens}.
For an input image $\obs_t$, its latent token sequence is denoted as $\tokens_t=(\token_t^1,\token_t^2,\cdots,\token_t^K)$.
To map a token sequence back to the input space, we first retrieve the embedding for each token and obtain a sequence $(\decLatent_t^1,\decLatent_t^2,\cdots,\decLatent_t^K)$ where $\decLatent_t^k = \tknEmb_{z_t^k}$.
Then, inverse to the encoding process, the decoder is responsible for mapping this sequence to a reconstructed observation $\hat{\obs}_t$.

The tokenizer is trained on frames sampled uniformly from the replay buffer. 
Its optimization objective, architecture, and other details are deferred to Appendix \ref{sec:tokenizer-learining-details}.

\paragraph{$\WM$ - World Model}

At the core of a world model is the component that captures the dynamics of the environment and makes predictions based on historical observations.
Here, $\WM$ is learned entirely in the latent token space, modeling the following distributions at each step $t$:
\begin{align}
\label{eq:wm-transitions-distributions}
    \text{Transition:}\quad & p(\hat{\tokens}_{t+1} \vert \tokens_{1}, \action_{1}, \ldots, \tokens_{t}, \action_{t} ), \\
    \label{eq:wm-reward-distributions}
    \text{Reward:}\quad & p(\hat{\reward}_{t} \vert \tokens_{1}, \action_{1}, \ldots, \tokens_{t}, \action_{t}), \\
    \label{eq:wm-termination-distributions}
    \text{Termination:}\quad & p(\hat{\doneSgnl}_{t} \vert \tokens_{1}, \action_{1}, \ldots, \tokens_{t}, \action_{t}).
\end{align}


To map observation tokens to embedding vectors, $\WM$ uses the code vectors $\tknzrCodebook$ learned by the tokenizer $\Tokenizer$.
Note that $\tknzrCodebook$ is not updated by $\WM$.
In addition, $\WM$ maintains dedicated embedding tables for mapping actions and special tokens (detailed in Section \ref{subsec:method-infer}) to continuous vectors.


\paragraph{$\Controller$ - Controller}

REM's actor-critic controller $\Controller$ is trained to maximize return entirely in imagination ~\cite{Kaiser2020Model,hafner2021dreamerv2,micheli2022transformers}.
$\Controller$ comprises of a policy network $\policy$ and a value function estimator $V^{\policy}$, and operates on latent tokens and their embeddings. 
In each optimization step, $\WM$ and $\Controller$ are initialized with a short trajectory segment sampled from the replay buffer.
Subsequently, the agent interacts with the world model for $\horizon$ steps.
At each step $t$, the agent plays an action sampled from its policy $\policy(\action_{t} \vert \tokens_{1}, \action_{1}, \ldots, \tokens_{t-1}, \action_{t-1}, \tokens_{t})$.
The world model evolves accordingly, generating $\hat{\reward}_t$, $\hat{\doneSgnl}_t$, and $\hat{\tokens}_{t+1}$ by sampling from the appropriate distributions (Eqn. ~(\ref{eq:wm-transitions-distributions}-\ref{eq:wm-termination-distributions})). 
The resulting trajectories are then used to train the agent. 
Following \cite{micheli2022transformers}, we adopted the actor-critic objectives of DreamerV2 \cite{hafner2021dreamerv2}.
We leave the full details of its architecture and optimization to Appendix \ref{sec:actor-critic-details}.

\begin{figure}[t]
    \centering
    \includegraphics[width=\linewidth]{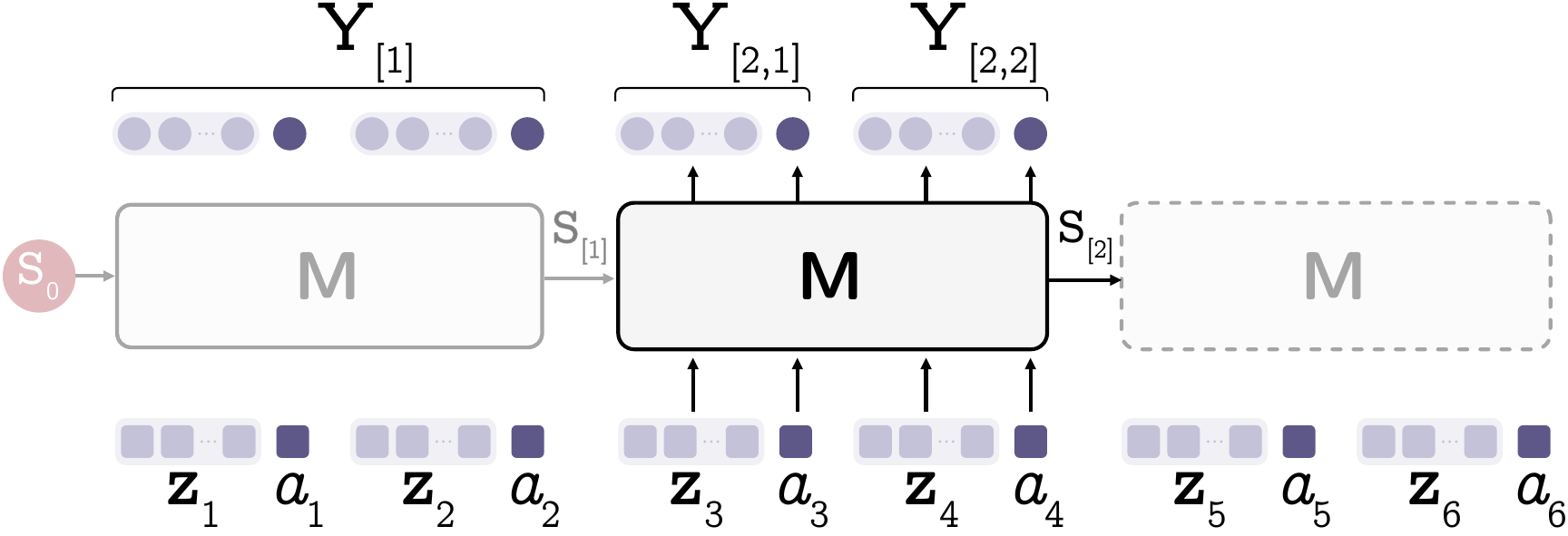}
    \caption{The ``chunkwise" computation mode. Long sequences can be split into smaller ``chunks" for enhanced training efficiency. Previous chunks are summarized by the recurrent state $\retnetState$.  Blue squares represents tokens, while circles denote output vectors. Crucially, RetNet's chunkwise mode does not natively support both a batched generation of tokens at imagination and an efficient world model training.
    These are achieved by our POP extension.
    }
    \label{fig:retention-chunkwise}
\end{figure}

\subsection{Retention Preliminaries}
\label{sec:retention-preliminaries}
Similar to Transformers \cite{Vaswani2017Attention}, a RetNet model \cite{sun2023retentive} consists of a stack of layers, where each layer contains a multi-head Attention-like mechanism, called Retention, followed by a fully-connected network. 
A unique characteristic of the Retention mechanism is that it has a dual form of recurrence and parallelism, called ``chunkwise", for improved efficiency when handling long sequences. 
This form allows to split such sequences into smaller ``chunks", where a parallel computation takes place within chunks and a sequential recurrent form is used between chunks, as shown in Figure~\ref{fig:retention-chunkwise}.
The information from previous chunks is summarized by a recurrent state $\retnetState\in\mathbb{R}^{d\times d}$ maintained by the Retention mechanism.

Formally, consider a sequence of tokens $(x_1, x_2, \cdots, x_m)$.
In our RL context, this sequence is a token trajectory composed of observation-action sub-sequences $(\token_t^1,\cdots,\token_t^K, a_t)$ we call \emph{blocks}.
As such trajectories are typically long, we split them into chunks of $\retnetChunkSize$ tokens, where $\retnetChunkSize = \blocksPerChunk (\tokensPerObs + 1)$ is a multiple of $\tokensPerObs + 1$ so that each chunk only contains complete blocks.
Here, the hyperparameter $\blocksPerChunk$ can be tuned according to the size of the models, the hardware, and other factors to maximize efficiency.
Let $\mathbf{X}=(\mathbf{x}_1, \mathbf{x}_2, \cdots, \mathbf{x}_m) \in \mathbb{R}^{m\times d}$ be the $d$-dimensional token embedding vectors.
The Retention output $\retnetY_{[i]} = \retentionOp{\retnetInput_{[i]}, \retnetState_{[i-1]}, i}$ of the $i$-th chunk is given by
\begin{equation}
 \label{eq:retnet-chunkwise-output}
     \retnetY_{[i]} = \left( \retnetQ_{[i]} \retnetK^{\Tr}_{[i]} \odot \retnetD \right) \retnetV_{[i]}   +    (\retnetQ_{[i]} \retnetState_{[i-1]}) \odot \boldsymbol{\xi},
\end{equation}
where the bracketed subscript $[i]$ is used to index the $i$-th chunk, $\retnetQ = \left( \retnetInput \retnetW_{Q} \right) \odot \retnetPos$, $\retnetK =  \left( \retnetInput \retnetW_{K} \right) \odot \retnetPosC$, $\retnetV = \retnetInput \retnetW_{V}$, and $\boldsymbol{\xi}\in\mathbb{R}^{B\times d}$ is a matrix with $\boldsymbol{\xi}_{ij} = \retnetEta^{i+1}$.
Here, 
  $\retnetW_{Q},\retnetW_{K},\retnetW_{V}\in\mathbb{R}^{d\times d}$ are learnable weights, $\retnetEta$ is an exponential decay factor, the matrix $\retnetD\in\mathbb{R}^{\retnetChunkSize \times \retnetChunkSize}$ combines an auto-regressive mask with the temporal decay factor $\retnetEta$, and the matrices $\retnetPos,\retnetPosC\in\mathbb{C}^{m\times d}$ are for relative position embedding (see Appendix \ref{sec:retentive-nets-details}).
Note the chunk index $i$ argument of the Retention operator, which controls positional embedding information through $\retnetPos$.
The chunkwise update rule of the recurrent state is given by
\begin{equation}
\label{eq:retnet-chunkwise-state-update}
    \retnetState_{[i]} = (\retnetK_{[i]} \odot \boldsymbol{\zeta})^{\Tr} \retnetV_{[i]} + \retnetEta^{\retnetChunkSize} \retnetState_{[i-1]}
\end{equation}
where $\retnetState_{[0]} = \retnetState_{0} = 0$, and $\boldsymbol{\zeta}\in\mathbb{R}^{B\times d}$ is a matrix with $\boldsymbol{\zeta}_{ij} = \retnetEta^{\retnetChunkSize - i - 1}$.
On the right hand side of Equations \ref{eq:retnet-chunkwise-output} and \ref{eq:retnet-chunkwise-state-update}, the first term corresponds to the computation within the chunk while the second term incorporates the information from previous chunks, encapsulated by the recurrent state.
Further details about the RetNet architecture are deferred to Appendix \ref{sec:retentive-nets-details}.

\begin{figure}[t]
    \centering
    \includegraphics[width=\linewidth]{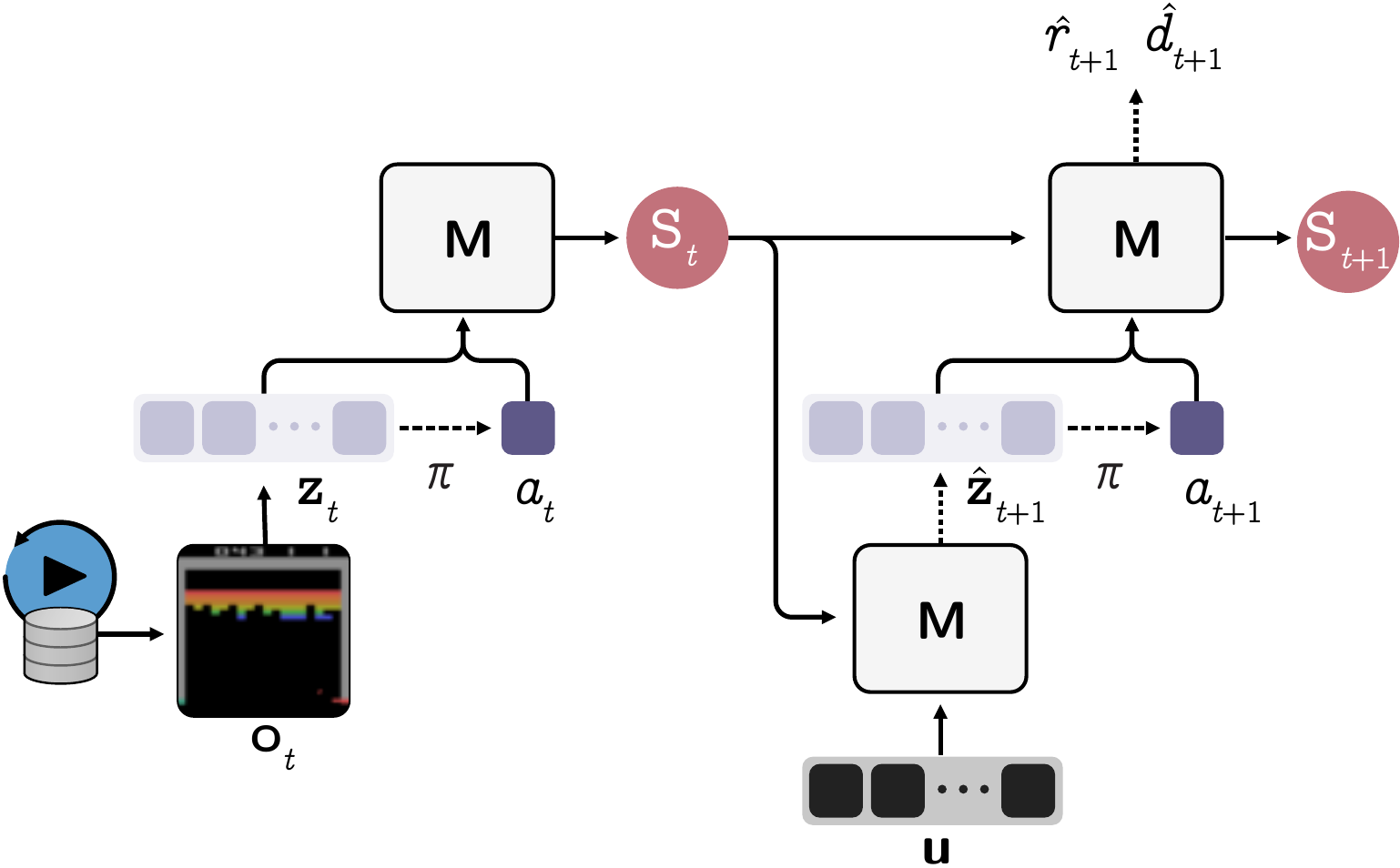}
    \caption{A single imagination step. Starting from a recurrent state $\retnetState_{t}$, initially obtained from real experience, $\WM$ computes all next-observation tokens $\hat{\tokens}_{t+1}$ in parallel using the prediction tokens $\predTokens$ as inputs. Then, the agent observes $\hat{\tokens}_{t+1}$ and picks an action $a_{t+1}$. Finally, $\WM$ takes $\retnetState_{t}, \hat{\tokens}_{t+1},$ and $\action_{t+1}$ and outputs $\retnetState_{t+1}, \hat{r}_{t+1}, \hat{d}_{t+1}$. Dashed arrows emphasize sampling operations. }
    \label{fig:imagination}
\end{figure}

\subsection{World Model Imagination}
\label{subsec:method-infer}

As the agent's training relies entirely on world model imagination, the efficiency of the trajectory generation is critical.
During imagination, predicting $\hat{\tokens}_{t+1}$ constitutes the primary non-trivial component and consumes the majority of processing time.
In IRIS, the prediction of $\hat{\tokens}_{t+1}$ unfolds sequentially, as the model is limited to predicting only one token ahead at each step.
This limitation arises since the identity of the next token, which remains unknown at the current step, is necessary for the prediction of later tokens.
Thus, generating $\horizon$ observations costs $\tokensPerObs \horizon$ sequential world model calls.
This leads to poor GPU utilization and long computation time.



To overcome this bottleneck, POP maintains a set of $\tokensPerObs$ dedicated prediction tokens $\predTokens = ( \predToken_{1},\ldots , \predToken_{\tokensPerObs})$ together with their corresponding embeddings $\predEmb \in \mathbb{R}^{\tokensPerObs \times \retnetDmodel}$.
To generate $\hat{\tokens}_{t+1}$ in one pass, POP simply computes the RetNet outputs starting from $\retnetState_{t}$ using $\predTokens$ as its input sequence, as illustrated in Figure~\ref{fig:imagination}.
Note that at imagination, the chunk size is limited to a single block, i.e., to $\tokensPerObs + 1$.
Here, the notation $\retnetState_{t}$ refers to the state that summarizes the first $t$ observation-action blocks.
To obtain $\retnetState_{t}$, we use RetNet's chunkwise forward to summarize an initial context segment of blocks sampled from the replay buffer.
Essentially, for every $t$, POP models the following distribution for next observation prediction:
\begin{equation*}
    p(\hat{\tokens}_{t+1} \vert \tokens_{1}, \action_{1}, \ldots, \tokens_{t}, \action_{t}, \predTokens)
\end{equation*}
with
\begin{equation*}
    p(\hat{\tokens}_{t+1}^{k} \vert \tokens_{1}, \action_{1}, \ldots, \tokens_{t}, \action_{t}, \predTokens_{\leq k}).
\end{equation*}
It is worth noting that the tokens $\predTokens$ are only intended for observation token predictions and are never employed in the update of the recurrent state.

This approach effectively reduces the total number of world model calls during imagination from $\tokensPerObs \horizon$ to $2\horizon$, eliminating the dependency on the number of observation tokens $\tokensPerObs$.
In fact, POP provides an additional generation mode that further reduces the number of sequential calls to $\horizon$.
We defer the details on this mode to Appendix \ref{sec:wm-details}.
Also, by using a recurrent state that summarizes long history sequences, POP improves efficiency further, as the per-token prediction cost reduces.
Effectively, POP offers improved scalability at the expense of a higher overall computational cost ($(2 \tokensPerObs + 1) \horizon$ compared to $(\tokensPerObs + 1) \horizon$).
Our approach add to existing evidence suggesting that enhanced scalability is often favorable, even at the expense of additional computational costs, with Transformers \cite{Vaswani2017Attention} serving as a prominent example.

\begin{figure}[t]
    \centering
    \includegraphics[width=\linewidth]{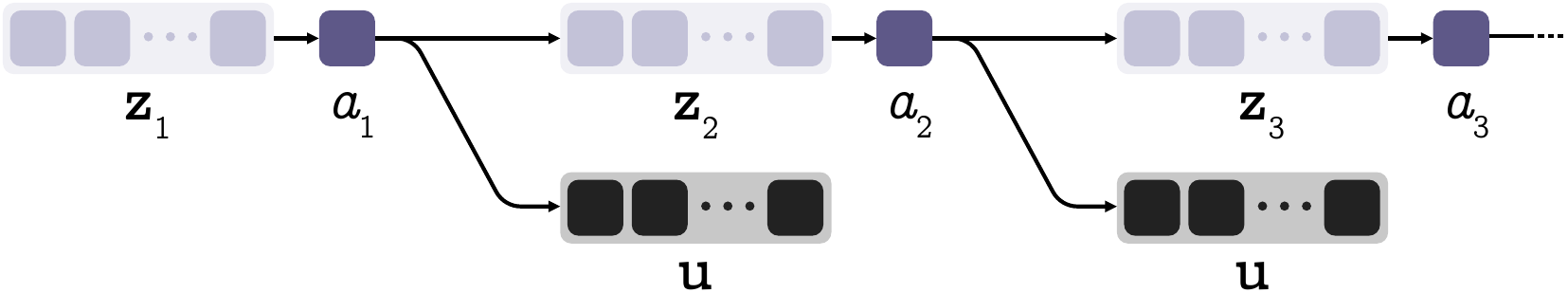}
    \caption{When appending $\predTokens$ after an observation-action block, the sequence is no longer a prefix of the observation-action token trajectory. Thus, the recurrent state only summarizes observation and action tokens (top trajectory).}
    \label{fig:training-sequence-prefix-problem}
\end{figure}




\subsection{World Model Training}
\label{subsec:method-train}

While applying POP during imagination is fairly straightforward, it requires modification of the training data.
Consider an input trajectory segment $(\tokens_{1}, \action_{1}, \ldots, \tokens_{T}, \action_{T})$ sampled from the replay buffer.
To make meaningful observation predictions at imagination, the model should be trained to predict $\tokens_{t}$ given $(\tokens_{1}, \action_{1}, \ldots, \tokens_{t-1}, \action_{t-1}, \predTokens)$, for each time step $t$ of every input segment.
Hence, for every $t$, the input sequence should contain $\predTokens$ at block $t$. 
However, replacing $\tokens_{t}$ with $\predTokens$ in the original sequence is inadequate, as the prediction of future observations, rewards, and termination signals depends on $\tokens_{t}$.
Thus, the standard approach of computing all outputs from the same input sequence is not viable, as in this case these two requirements contradict each other (Figure~\ref{fig:training-sequence-prefix-problem}).
The challenge then lies in devising an efficient method for computing the outputs for all time steps in parallel.
To tackle this challenge, we first note that each trajectory prefix can be summarized into a single recurrent state.
For example, for the first input chunk 
$(\tokens_1, \action_1, \ldots, \tokens_{\blocksPerChunk}, \action_{\blocksPerChunk})$,
$(\tokens_1, a_1)$ can be summarized into $\retnetState_{[1, 1]}$ and $(\tokens_1, a_1,\tokens_2, a_2)$ can be summarized into $\retnetState_{[1, 2]}$.
Here,  we use the subscript $[i, j]$ to conveniently refer to the $j$-th block within the $i$-th chunk (this notation is demonstrated in Figure \ref{fig:retention-chunkwise}), with $\retnetState_{[i, 0]} = \retnetState_{[i-1]}$ and $\retnetState_{[i, \blocksPerChunk]} = \retnetState_{[i]}$.
Thus, our plan is to first compute all states $\retnetState_{[i, 1]}, \ldots, \retnetState_{[i, \blocksPerChunk]}$ in parallel, and then predict all next observations from all $(\retnetState_{[i, j]}, \predTokens)$ tuples.

\begin{figure}[t]
    \centering
    \includegraphics[width=\linewidth]{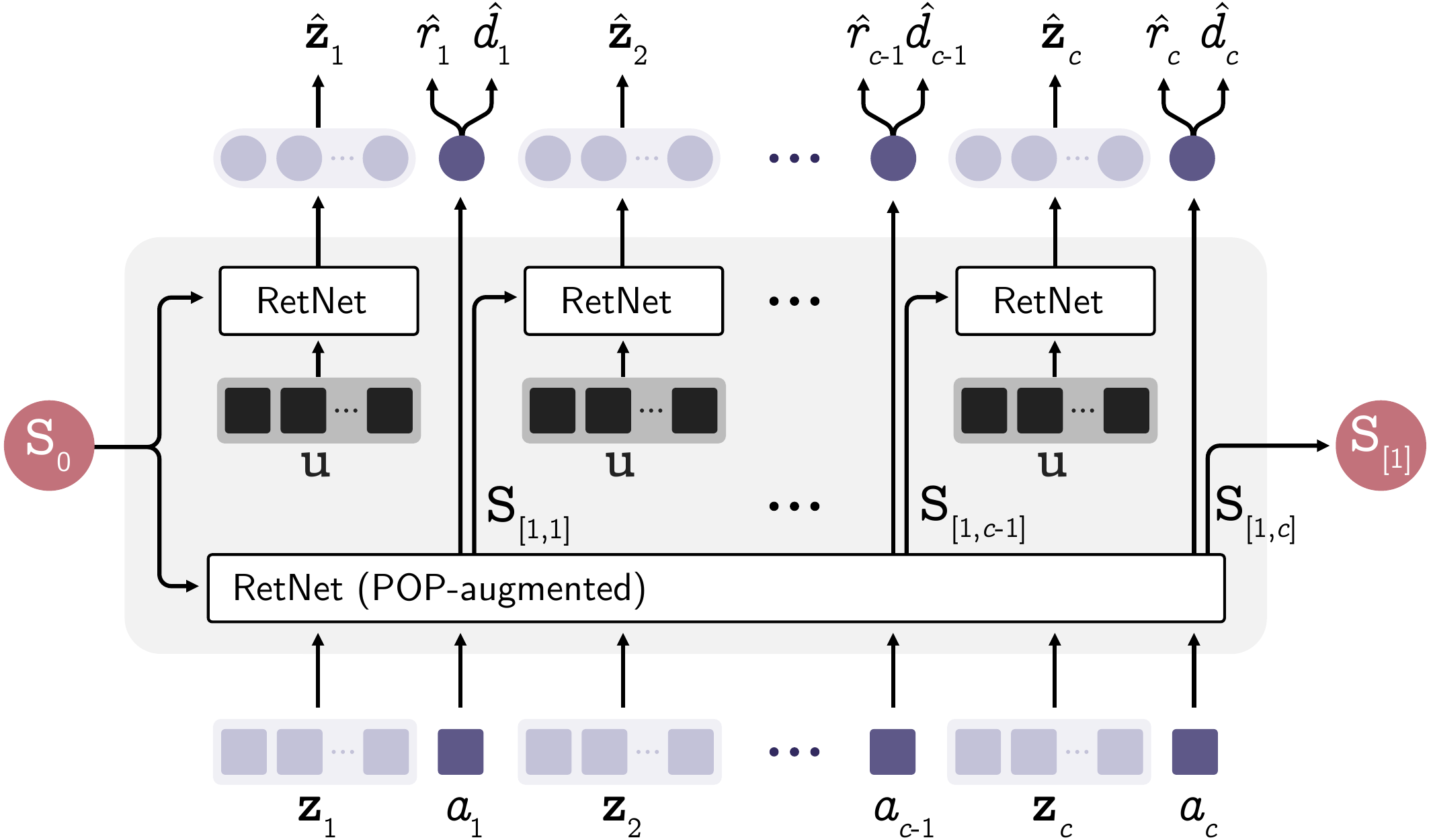}
    \caption{An illustration of the POP chunkwise forward algorithm (Alg. \ref{alg:pop-retnet-fwd-training} and \ref{alg:pop-layer-fwd-training}) for a single-layer model. During training, $\WM$ computes the outputs of $\blocksPerChunk$ observation-action blocks in parallel. Blue squares represent token inputs, while the corresponding RetNet outputs are denoted by circles.
    Each RetNet block represents a forward call to the same RetNet model. 
    The bottom RetNet call uses our POP extension for computing the additional recurrent states at the end of every observation-action block (Alg. \ref{alg:pop-layer-fwd-training}, lines 2-7). 
    The top row of RetNet calls are batch-computed in parallel (Alg. \ref{alg:pop-layer-fwd-training}, line 8). 
    Finally, the output combines the observation token outputs produced by the top RetNet call with the rewards and termination outputs computed by the bottom one (Alg. \ref{alg:pop-retnet-fwd-training}, lines 7-9).}
    \label{fig:wm-training}
\end{figure}

To compute all recurrent states $\retnetState_{[i, j]}$ in parallel, a two-step computation is carried.
First, intermediate states $\Tilde{\retnetState}_{[i, j]}$ are computed in parallel for all $j$ with
\begin{equation}
\label{eq:pop-pseudo-state}
    \Tilde{\retnetState}_{[i, j]} = \left( \retnetK_{[i, j]} \odot \boldsymbol{\zeta} \right)^{\Tr} \retnetV_{[i, j]},
\end{equation}
where $\boldsymbol{\zeta}\in\mathbb{R}^{(K+1)\times d}$ is a matrix with $\boldsymbol{\zeta}_{ij}=\eta^{K-i}$.
Then, each recurrent state is computed sequentially by
\begin{equation}
\label{eq:pop-state-from-pseudo}
    \retnetState_{[i, j]} = \Tilde{\retnetState}_{[i, j]} + \retnetEta^{\tokensPerObs + 1} \retnetState_{[i, j-1]}.
\end{equation}
As the majority of the computational burden lies in the first step, the sequential computation in the second step has minimal impact on the overall speedup.

Once we have all states ready, the output of $(\retnetState_{[i, j]}, \predTokens)$ for all $1 \leq j \leq \blocksPerChunk$ is computed in parallel.
Here, we stress that the existing Retention mechanism can only perform batched input computation with recurrent states $\retnetState_{t}$ of the same time step $t$.
This is due to the shared positional embedding information applied to every input sequence in the batch.
To overcome this, we devise a mechanism which extends RetNet to support the batched computation of the $(\retnetState_{[i, j]}, \predTokens)$ tuples, while applying the appropriate positional encoding information.
A pseudo code of our novel POP extension of RetNet is given in Algorithms \ref{alg:pop-retnet-fwd-training} and $\ref{alg:pop-layer-fwd-training}$.
The latter presents the core of the mechanism (described above), while the former describes the higher level layer-by-layer computation with a final aggregation for combining the produced outputs.
Figure \ref{fig:wm-training} illustrates a simplified example of the POP Forward mechanism (Algorithms \ref{alg:pop-retnet-fwd-training} and \ref{alg:pop-layer-fwd-training}) for a single-layer model.
For brevity, our pseudo code and illustrations only considers Retention layers, omitting other modules of RetNet (Appendix \ref{sec:retentive-nets-details}).

\begin{algorithm}[tb]
   \caption{RetNet POP Chunkwise Forward}
   \label{alg:pop-retnet-fwd-training}
\begin{algorithmic}[1]
   \STATE {\bfseries Input:} chunk size $1 \leq \blocksPerChunk \leq \horizon$, 
   token embeddings $\retnetInput_{[i]}$ of chunk $i$,
   per-layer recurrent states  $\{\retnetState_{[i-1]}^{l} \}_{l=1}^{L}$.

    \STATE Initialize $\popOutsA^{0}_{[i]} \leftarrow \retnetInput$

    \STATE Initialize $\popOutsB^{0}_{[i, 1]}, \ldots, \popOutsB^{0}_{[i, \blocksPerChunk]} \leftarrow \tknzrCodebook_\predTokens, \ldots, \tknzrCodebook_\predTokens$
    
    \FOR{$l=1$ {\bfseries to} $L$}

    \STATE $\popOutsA^{l}_{[i]}, \popOutsB^{l}_{[i]}, \retnetState_{[i]}^{l} \leftarrow \text{POPLayer}(\popOutsA^{l-1}_{[i]}, \popOutsB^{l-1}_{[i]}, \retnetState_{[i-1]}^{l}, i)$

    \ENDFOR

    \FOR{$j=1$ {\bfseries to} $\blocksPerChunk$}
    \STATE $\retnetY_{[i, j]} \leftarrow \text{Concat}(\popOutsB^{L}_{[i, j]}, \popOutsA^{L}_{[i, j, \tokensPerObs + 1]}) $ 
    
    \ENDFOR

    \STATE {\bfseries Return}  $\retnetY, \{ \retnetState_{[i]}^{l} \}_{l=1}^{L}$
\end{algorithmic}
\end{algorithm}

\begin{algorithm}[tb]
   \caption{POPLayer Chunkise Forward}
   \label{alg:pop-layer-fwd-training}
\begin{algorithmic}[1]
   \STATE {\bfseries Input:} Chunk latents $\popOutsA_{[i]}$, observation prediction latents $\popOutsB_{[i]}$, recurrent state $\retnetState_{[i-1]}$, chunk index $i$.


    \STATE $\popOutsA_{[i]} \leftarrow \retentionOp{\popOutsA_{[i]}, \retnetState_{[i-1]}, i}$ (Eqn. \ref{eq:retnet-chunkwise-output}) 
    
    \STATE Compute $\Tilde{\retnetState}_{[i, 1]}, \ldots, \Tilde{\retnetState}_{[i, \blocksPerChunk]}$ in parallel (Eqn. \ref{eq:pop-pseudo-state})

    \FOR{$j=1$ {\bfseries to} $\blocksPerChunk$ }
   \STATE $\retnetState_{[i, j]} \leftarrow \Tilde{\retnetState}_{[i, j]} + \retnetEta^{\tokensPerObs + 1} \retnetState_{[i, j-1]}$ (Eqn. \ref{eq:pop-state-from-pseudo})
   \ENDFOR

   \STATE $\retnetState_{[i]} \leftarrow \retnetState_{[i, \blocksPerChunk]}$

    \STATE $\popOutsB_{[i, j]} \leftarrow  \retentionOp{\popOutsB_{[i, j]}, \retnetState_{[i, j-1]}, [i, j]}$ in parallel for $j=1,\ldots, \blocksPerChunk$ (Eqn. \ref{eq:retnet-chunkwise-output})

    \STATE {\bfseries Return}  $\popOutsA_{[i]}, \popOutsB_{[i]}, \retnetState_{[i]}$
\end{algorithmic}
\end{algorithm}

To train the world model, trajectory segments of $\horizon$ steps from past experience are uniformly sampled from the replay buffer and translated into token sequences.
These sequences are processed in chunks of $\blocksPerChunk$ observation-action blocks to produce the modeled distributions, as depicted in Figure \ref{fig:wm-training}.
Optimization is carried by minimizing the cross-entropy loss of the transitions and termination outputs, and the appropriate loss of the reward outputs, depending on the task.
For continuous rewards, the mean-squared error loss is used while for discrete ones cross-entropy is used instead.

\begin{table*}[t]
\caption{Mean agent returns on the 26 games of the Atari 100k benchmark followed by averaged human-normalized performance metrics. Each game score is computed as the average of 5 runs with different seeds, where the score of each run is computed as the average over 100 episodes sampled at the end of training. Bold face and underscores mark the highest score among token-based methods and among all baselines, respectively.}
\label{table:main-results}
\vskip 0.15in
\begin{center}
\begin{small}
\begin{tabular}{lcc cccc cr}
\toprule

\multicolumn{3}{c}{} & \multicolumn{4}{c}{Non-Token-Based} & \multicolumn{2}{c}{Token-Based} \\
\cmidrule(lr){4-7} \cmidrule(lr){8-9} 

Game                 &  Random    &  Human     &  SimPLe    &  DreamerV3             &  TWM                   &  STORM                 &  IRIS               &  \textsc{REM} (ours)          \\
\midrule
Alien                &  227.8     &  7127.7    &  616.9     &  959.4                 &  674.6                 &  \underline{983.6}     &  420.0              &  \textbf{607.2}               \\
Amidar               &  5.8       &  1719.5    &  74.3      &  139.1                 &  121.8                 &  \underline{204.8}     &  \textbf{143.0}     &  95.3                         \\
Assault              &  222.4     &  742.0     &  527.2     &  705.6                 &  682.6                 &  801.0                 &  1524.4             &  \underline{\textbf{1764.2}}  \\
Asterix              &  210.0     &  8503.3    &  1128.3    &  932.5                 &  1116.6                &  1028.0                &  853.6              &  \underline{\textbf{1637.5}}  \\
BankHeist            &  14.2      &  753.1     &  34.2      &  \underline{648.7}     &  466.7                 &  641.2                 &  \textbf{53.1}      &  19.2                         \\
BattleZone           &  2360.0    &  37187.5   &  4031.2    &  12250.0               &  5068.0                &  \underline{13540.0}   &  \textbf{13074.0}   &  11826.0                      \\
Boxing               &  0.1       &  12.1      &  7.8       &  78.0                  &  77.5                  &  79.7                  &  70.1               &  \underline{\textbf{87.5}}    \\
Breakout             &  1.7       &  30.5      &  16.4      &  31.1                  &  20.0                  &  15.9                  &  83.7               &  \underline{\textbf{90.7}}    \\
ChopperCommand       &  811.0     &  7387.8    &  979.4     &  410.0                 &  1697.4                &  1888.0                &  1565.0             &  \underline{\textbf{2561.2}}  \\
CrazyClimber         &  10780.5   &  35829.4   &  62583.6   &  \underline{97190.0}   &  71820.4               &  66776.0               &  59324.2            &  \textbf{76547.6}             \\
DemonAttack          &  152.1     &  1971.0    &  208.1     &  303.3                 &  350.2                 &  164.6                 &  2034.4             &  \underline{\textbf{5738.6}}  \\
Freeway              &  0.0       &  29.6      &  16.7      &  0.0                   &  24.3                  &  0.0                   &  31.1               &  \underline{\textbf{32.3}}    \\
Frostbite            &  65.2      &  4334.7    &  236.9     &  909.4                 &  \underline{1475.6}    &  1316.0                &  \textbf{259.1}     &  240.5                        \\
Gopher               &  257.6     &  2412.5    &  596.8     &  3730.0                &  1674.8                &  \underline{8239.6}    &  2236.1             &  \textbf{5452.4}              \\
Hero                 &  1027.0    &  30826.4   &  2656.6    &  \underline{11160.5}   &  7254.0                &  11044.3               &  \textbf{7037.4}    &  6484.8                       \\
Jamesbond            &  29.0      &  302.8     &  100.5     &  444.6                 &  362.4                 &  \underline{509.0}     &  \textbf{462.7}     &  391.2                        \\
Kangaroo             &  52.0      &  3035.0    &  51.2      &  4098.3                &  1240.0                &  \underline{4208.0}    &  \textbf{838.2}     &  467.6                        \\
Krull                &  1598.0    &  2665.5    &  2204.8    &  7781.5                &  6349.2                &  \underline{8412.6}    &  \textbf{6616.4}    &  4017.7                       \\
KungFuMaster         &  258.5     &  22736.3   &  14862.5   &  21420.0               &  24554.6               &  \underline{26182.0}   &  21759.8            &  \textbf{25172.2}             \\
MsPacman             &  307.3     &  6951.6    &  1480.0    &  1326.9                &  1588.4                &  \underline{2673.5}    &  \textbf{999.1}     &  962.5                        \\
Pong                 &  -20.7     &  14.6      &  12.8      &  18.4                  &  \underline{18.8}      &  11.3                  &  14.6               &  \textbf{18.0}                \\
PrivateEye           &  24.9      &  69571.3   &  35.0      &  881.6                 &  86.6                  &  \underline{7781.0}    &  \textbf{100.0}     &  99.6                         \\
Qbert                &  163.9     &  13455.0   &  1288.8    &  3405.1                &  3330.8                &  \underline{4522.5}    &  \textbf{745.7}     &  743.0                        \\
RoadRunner           &  11.5      &  7845.0    &  5640.6    &  15565.0               &  9109.0                &  \underline{17564.0}   &  9614.6             &  \textbf{14060.2}             \\
Seaquest             &  68.4      &  42054.7   &  683.3     &  618.0                 &  774.4                 &  525.2                 &  661.3              &  \underline{\textbf{1036.7}}  \\
UpNDown              &  533.4     &  11693.2   &  3350.3    &  7567.1                &  \underline{15981.7}   &  7985.0                &  3546.2             &  \textbf{3757.6}              \\
\midrule
\#Superhuman (↑)     &  0         &  N/A       &  1         &  9                     &  8                     &  9                     &  10                 &  \underline{\textbf{12}}      \\
Mean (↑)             &  0.000     &  1.000     &  0.332     &  1.124                 &  0.956                 &  \underline{1.222}     &  1.046              &  \textbf{1.222}               \\
Median (↑)           &  0.000     &  1.000     &  0.134     &  0.485                 &  \underline{0.505}     &  0.425                 &  \textbf{0.289}     &  0.280                        \\
IQM (↑)              &  0.000     &  1.000     &  0.130     &  0.487                 &  0.459                 &  0.561                 &  0.501              &  \underline{\textbf{0.673}}   \\
Optimality Gap (↓)   &  1.000     &  0.000     &  0.729     &  0.510                 &  0.513                 &  \underline{0.472}     &  0.512              &  \textbf{0.482}               \\

\bottomrule
\end{tabular}
\end{small}
\end{center}
\vskip -0.1in
\end{table*}

\begin{figure*}[t]
    \centering
    \includegraphics[width=0.98\linewidth]{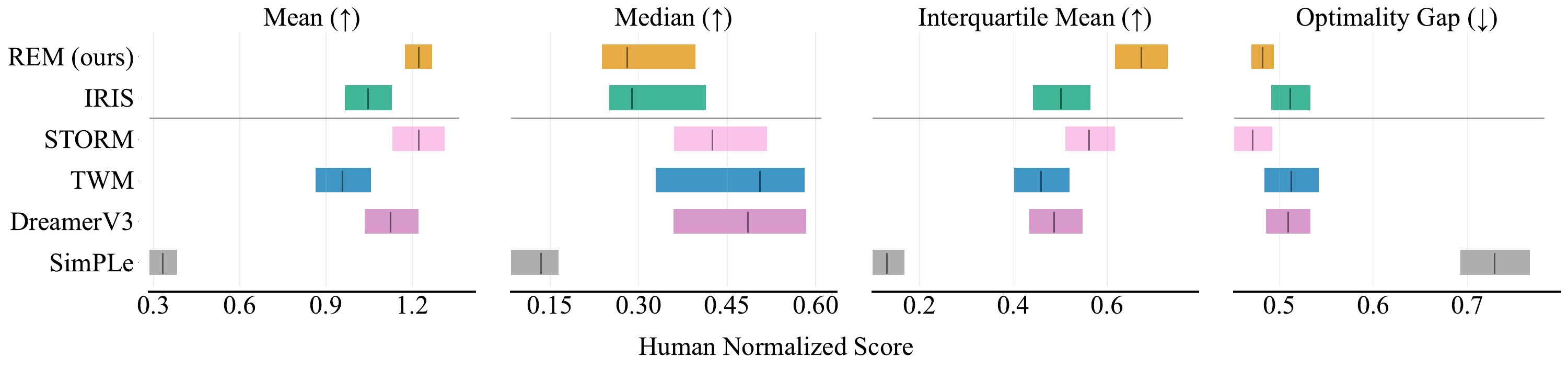}
    \caption{Atari 100K aggregated metrics with 95\% stratified bootstrap confidence intervals of the mean, median, and inter-quantile mean (IQM) human-normalized scores and optimality gap \cite{Agarwal2021rliable}. A line separates token-based methods from other baselines. }
    \label{fig:benchmark-aggregates}
\end{figure*}

\section{Experiments}
\label{sec:experiments}
We follow most prior works on world models and evaluate REM on the widely-recognized Atari 100K benchmark \cite{Kaiser2020Model} for sample-efficient reinforcement learning.
The Atari 100K benchmark considers a subset of 26 Atari games. 
For each game, the agent is limited to 100K interaction steps, corresponding to 400K game frames due to the standard frame-skip of 4.
In total, this amounts to roughly 2 hours of gameplay.
To put in perspective, the original Atari benchmark allows agent to collect 200M steps, that is, 500 times more experience.

\paragraph{Experimental Setup}

The full details of the architectures and hyper-parameters used in our experiments are presented in Appendix \ref{sec:models-and-hyperparameters}.
Notably, our tokenizer uses $\tokensPerObs=64$ (i.e., a grid of $8 \times 8$ latent tokens per observation), whereas IRIS uses only $\tokensPerObs=4\times 4 = 16$.
To ensure a meaningful comparison of the run times of REM and IRIS, REM's configuration was chosen so that the amount of computation carried by each component at each epoch remains (roughly) equivalent to that of the corresponding component in IRIS.
For benchmarking agents run times, we used a workstation with an Nvidia RTX 4090 GPU. 
The rest of our experiments were conducted on Nvidia V100 GPUs.


\paragraph{Baselines}
Since the contributions of this paper relate to token-based approaches, and to IRIS in particular, our evaluation focuses on token-based methods. 
To enrich our results, as well as to facilitate future research, we have included the following additional baselines: SimPLe \cite{Kaiser2020Model}, DreamerV3 \cite{hafner2023mastering}, TWM \cite{robine2023TWM}, and STORM \cite{zhang2023storm}.
In these approaches, observations are processed as a single sequence element by the world model.
Following prior works on world models, lookahead search methods such as MuZero \cite{Schrittwieser2020} and EfficientZero \cite{Ye2021EfficientZero} are not included as lookahead search operates on top of a world model. 
Here, our aim is to improve the world model component itself.

\subsection{Results}
\label{sec:exp-main-results}
\begin{figure}
    \centering
    \includegraphics[width=\linewidth]{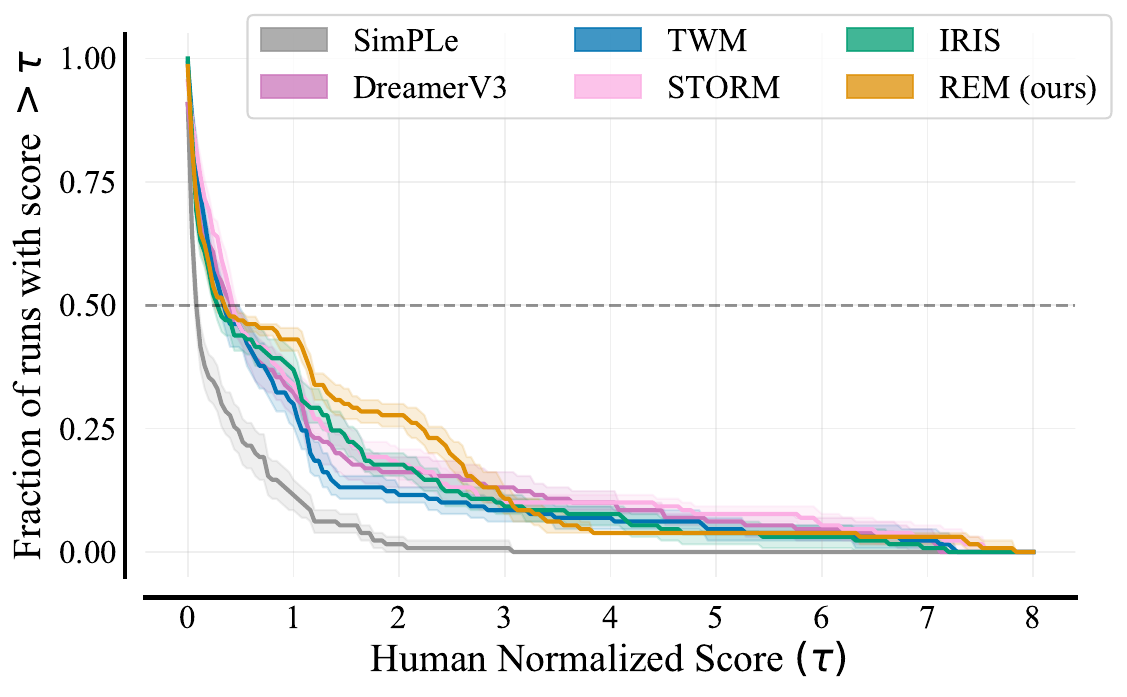}
    \caption{Performance profiles. For every human-normalized score value (x axis), each algorithm's curve shows the fraction of its runs with score grater than the given score value. The shaded area indicates pointwise 95\% confidence bands based on percentile bootstrap with stratified sampling  \cite{Agarwal2021rliable}.}
    \label{fig:benchmark-pref-prof}
\end{figure}

On Atari, it is standard to use human-normalized scores (HNS) \cite{mnih2015human}, calculated as $\frac{\text{agent\_score} \ - \  \text{random\_score} }{\text{human\_score} \  - \  \text{random\_score}}$, rather than raw game scores.
Here, the final score of each training run is computed as an average over 100 episodes collected at the end of training.
In the work of \cite{Agarwal2021rliable}, the authors found discrepancies between conclusions drawn from point estimate statistics such as mean and median and a more thorough statistical analysis that also considers the uncertainty in the results.
Adhering to their established protocol and utilizing their toolkit, we report the mean, median, and interquantile mean (IQM) human-normalized scores, and the optimality gap, with 95\% stratified bootstrap confidence intervals in Figure \ref{fig:benchmark-aggregates}.
Performance profiles are presented in Figure \ref{fig:benchmark-pref-prof}. 
Average scores of individual games are reported in \autoref{table:main-results}.

REM attains an IQM human normalized score of 0.673, outperforming all baselines.
Additionally, REM improves over IRIS on 3 out of the 4 metrics (i.e., mean, optimality gap, and IQM), while being comparable in terms of its median score.
Remarkably, REM achieves superhuman performance on 12 games, more than any other baseline (Table \ref{table:main-results}).
REM also exhibits state-of-the-art scores on several games, including Assault, Boxing, and Chopper Command.
These findings support our empirical claim that REM performs similarly or better than previous token-based approaches while running significantly faster.


\subsection{Ablation Studies}
\label{sec:exp-ablations-study}
To analyze the impact of different components of our approach on REM's performance, we conduct a series of ablation studies.
For each component, we compare the final algorithm to a version where the component of interest is disabled.
Due to computational resource constraints, the evaluation is performed on a subset of 8 games from the Atari 100K benchmark using 5 random seeds for each game.
This subset includes games with large score differences between IRIS and REM, as we are particularly interested in studying the impact of each component in these games. 
Concretely, this subset includes the games ``Assult", ``Asterix", ``Chopper Command", ``Crazy Climber", ``Demon Attack", ``Gopher", ``Krull", and ``Road Runner".
We performed ablation studies on the following aspects: the POP mechanism, the latent space architecture of $\Controller$ and its action inputs, the latent resolution of $\Tokenizer$, and the observation token embeddings used by $\WM$. 

\begin{figure}[t]
    \centering
    \includegraphics[width=\linewidth]{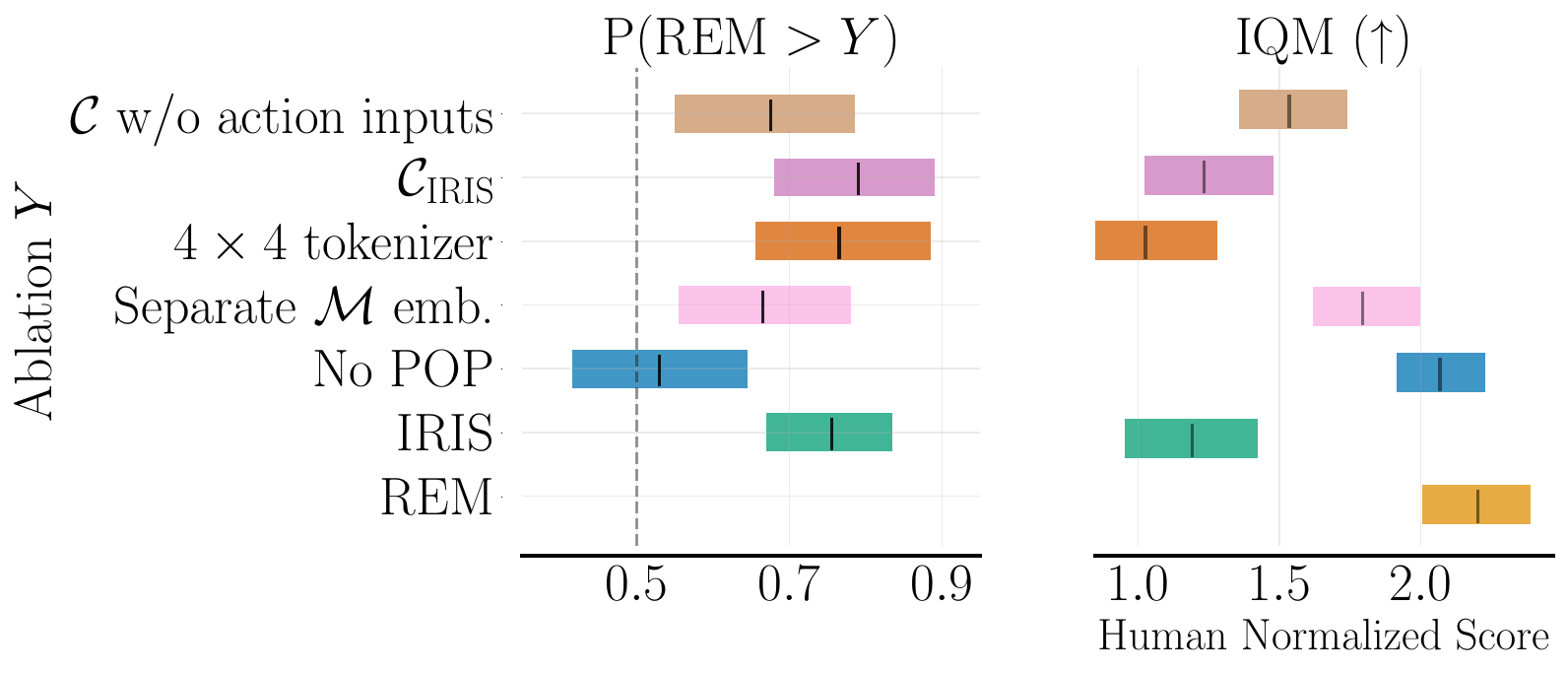}
    \caption{\textbf{Left}: The probability of improvement \cite{Agarwal2021rliable} shows the probability of REM outperforming each ablation on a randomly selected game from the subset of 8 games used for the ablation studies. \textbf{Right}: interquantile mean (IQM) human normalized score. Each band indicate a 95\% stratified bootstrap confidence interval.}
    \label{fig:ablations-summary}
\end{figure}

\begin{figure}[t]
    \centering
    \includegraphics[width=0.95\linewidth]{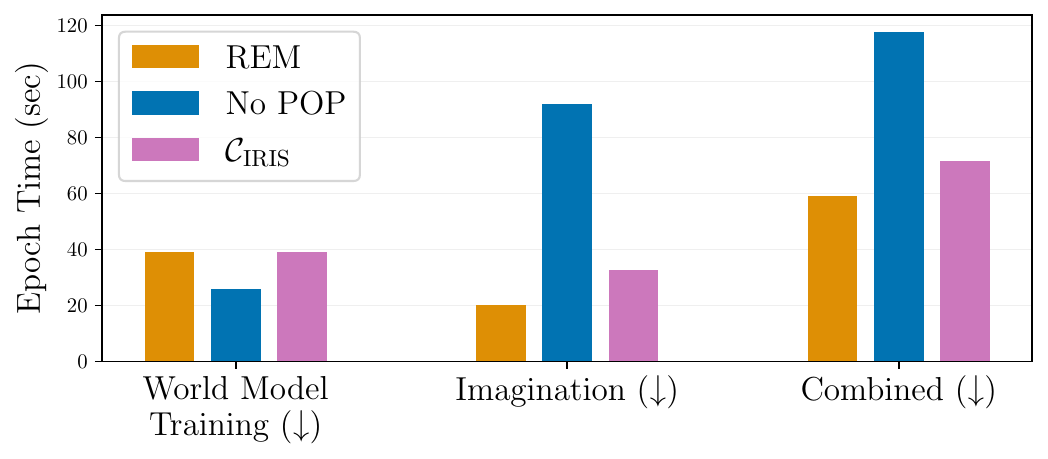}
    \caption{Epoch run time comparison between REM and two of its ablations: ``No POP", and $\Controller_{\text{IRIS}}$.}
    \label{fig:ablations-run-times-comparison}
\end{figure}

The probability of improvement \cite{Agarwal2021rliable} and IQM human-normalized scores are presented in Figure \ref{fig:ablations-summary}.
Figure \ref{fig:ablations-run-times-comparison} offers a comparison of the training times, juxtaposing REM with its efficiency-related ablations.


\paragraph{Analyzing POP}
To study the impact of POP on REM's performance, we replaced the POP-augmented RetNet of $\WM$ with a vanilla RetNet.
In this version, denoted as "No POP", the prediction of next observation tokens is performed sequentially token-by-token, as done in IRIS.

Our results suggest that POP retains the agent's performance (Figure \ref{fig:ablations-summary}) while significantly reducing the overall computation time (Figure \ref{fig:ablations-run-times-comparison}).
In Appendix \ref{sec:additional-results}, we provide additional results indicating that the world model's performance are also retained.
POP achieves lower total computation time by expediting the actor-critic learning phase, despite the increased computational cost implied by the observation prediction tokens.

\paragraph{Actor-Critic Architecture and Action Inputs}
For $\Controller$, we considered an incremental ablation. 
First, we replaced the architecture of REM's controller $\Controller$ with that of IRIS (denoted ``$\Controller_{\text{IRIS}}$").
In contrast to REM, this version processes fully reconstructed pixel frames and does not incorporate action inputs.
Formally, $\Controller_{\text{IRIS}}$ models $\policy(\action_{t} | \hat{\obs}_{\leq t}), V^{\policy}(\hat{\obs}_{\leq t})$.
In the second ablation, REM was modified so that only the action inputs of $\Controller$ were disabled.
This ablation corresponds to $\policy(\action_{t} | \hat{\tokens}_{\leq t}), V^{\policy}(\hat{\tokens}_{\leq t})$.

Our findings indicate that both the latent codes based architecture and the added action inputs contribute to the final performance of REM (Figure \ref{fig:ablations-summary}).
Additionally, the latent codes based architecture of $\Controller$ leads to reduced computational overhead and shorter actor-critic learning times (Figure \ref{fig:ablations-run-times-comparison}).

\paragraph{Tokenizer Resolution}
Here, we compare REM to a version with a reduced latent resolution of $4 \times 4$, similar to that of IRIS.
The results in Figure \ref{fig:ablations-summary} provides clear evidence that the latent resolution of the tokenizer has a significant impact on the agent's performance.
Our results demonstrates that POP enables REM to utilize higher latent resolutions while incurring shorter computation times than prior token-based approaches.

\paragraph{World Model Embeddings}
In REM, $\WM$ translates observation tokens to embedding vectors using the embedding table $\tknzrCodebook$ learned by $\Tokenizer$.
These embeddings encode the visual information as learned by $\Tokenizer$.
In contrast, IRIS maintains a separate embedding table learned by the world model for that purpose.
Here, the results in Figure \ref{fig:ablations-summary} provide empirical evidence indicating that leveraging this encoded visual information leads to improved performance.
In Appendix \ref{sec:additional-results}, we provide additional evidence suggesting that the world model's next-observation predictions are also improved.



\section{Related Work}
Model-based reinforcement learning (RL), with its roots in the tabular setting \cite{Sutton1991Dyna}, has been a focus of extensive research in recent decades.
The deep RL agent of \cite{ha2018worldmodels} leveraged an LSTM \cite{hochreiter1997lstm} sequence model with a VAE \cite{Kingma2014VAE} to model the dynamics in visual environments, demonstrating that successful policies can be learned entirely from simulated data. 
This approach, commonly known as world models, was later applied to Atari games \cite{Kaiser2020Model} with the PPO \cite{Schulman2017PPO} RL algorithm.
Later, a series of works \cite{Hafner2020Dreamerv1, hafner2021dreamerv2, hafner2023mastering} proposed the Dreamer algorithms, which are based on a recurrent state space model (RSSM) \cite{Hafner2019PlaNet} to model dynamics.
The latest DreamerV3 was evaluated on a variety of challenging environments, providing further evidence of the promising potential of world models.
In contrast to token-based approaches, where each token serves as a standalone sequence element, Dreamer encodes each frame as a vector of categorical variables, which are processed at once by the RSSM.

Following the success of the Transformer architecture \cite{Vaswani2017Attention} in language modeling \cite{brown2020LMsFewShotLearners}, and motivated by their favorable scaling properties compared to RNNs, Transformer were recently explored in RL \cite{Parisotto2020transformersRL, Chen2021DecisionTransfmr, reed2022aGATO, shridhar23aPercvrTsfmr}.
World model approaches also adopted the Transformer architecture.
\cite{micheli2022transformers} blazed the trail for token-based world models with IRIS, representing agent trajectories as language-like sequences.
By treating each observation as a sequence, its Transformer-based world model gains an explicit sub-observation attention resolution.
Despite IRIS's high performance, its imagination bottleneck results in a substantial disadvantage.

In addition to IRIS, non-token-based world models driven by Transformers were proposed.
TWM \cite{robine2023TWM} utilizes the Transformer-XL architecture \cite{Dai2020TsfmrXL} and a non-uniform data sampling. 
STORM \cite{zhang2023storm} proposes an efficient Transformer based world model agent which sets state-of-the-art result for the Atari 100K benchmark.
STORM has a significantly smaller 2-layer Transformer compared to the 10-layer models of TWM and IRIS, demonstrating drastically reduced training times and improved agent performance.




\section{Conclusions}

In this work, we presented a novel parallel observation prediction (POP) mechanism augmenting Retention networks with a dedicated forward mode to improve the efficiency of token-based world models (TBWMs).
POP effectively solves the imagination bottleneck of TBWMs and enables them to deal with longer observation sequences.
Additionally, we introduced REM, a TBWM agent equipped with POP.
REM is the first world model agent driven by the RetNet architecture.
Empirically, we demonstrated the superiority of REM over prior TBWMs on the Atari 100K benchmark, rendering REM competitive with the state-of-the-art, both in terms of agent performance and overall run time.

Our work opens up many promising avenues for future research by making TBWMs practical and cost-efficient.
One such direction could be to explore a modification of REM where the recurrent state of the world model summarizes the entire history of the agent.
Similarly, a history-preserving RetNet architecture should be considered for the controller as well.  
Another promising avenue would be to leverage the independent optimization of the tokenizer to enable REM to use pretrained visual perception models in environments where visual data is abundant, for example, the real world.
Such perceptual models could be trained at scale, and allow REM to store only compressed observations in its replay buffer, further improving its efficiency.
Lastly, token-based methods for video generation tasks can benefit from using the POP mechanism for generating entire frames in parallel conditioned on the past context.
We believe that this is an exciting avenue to explore with a potentially high impact.

\section*{Acknowledgements}
This project has received funding from the European Union’s Horizon Europe Programme under grant agreement No. 101070568.

\section*{Impact Statement}
This paper presents work whose goal is to advance the field of Machine Learning. There are many potential societal consequences of our work, none of which we feel must be specifically highlighted here.


\bibliographystyle{icml2024}

\newpage
\appendix
\onecolumn
\section{Appendix}

\subsection{Models and Hyperparameters}
\label{sec:models-and-hyperparameters}

Tables \ref{table:shared-hyperparams} and \ref{table:per-component-hyperparams} detail hyperparameters of the optimization and environment, as well as hyperparameters shared by multiple components.

\begin{table}[h]
\caption{Shared Hyperparameters}
\label{table:shared-hyperparams}
\vskip 0.15in
\begin{center}
\begin{small}
\begin{tabular}{lccr}
\toprule
Description & Symbol & Value \\
\midrule
Horizon & \horizon & 10 \\
Tokens per observation & \tokensPerObs & 64 \\
Tokenizer vocabulary size & \tknzrVocabSize & 512 \\
\midrule
Epochs & - & 600 \\
Experience collection epochs & - & 500 \\
Environment steps per epoch & - & 200 \\
Collection epsilon-greedy & - & 0.01 \\
Eval sampling temperature & - & 0.5 \\
Optimizer & - & AdamW \\
AdamW $\beta_1$ & - & 0.9 \\
AdamW $\beta_2$ & - & 0.999 \\

\midrule

Frame resolution & - & $64 \times 64$ \\
Frame Skip & - & 4 \\
Max no-ops (train, test) & - & (30, 1) \\
Max episode steps (train, test) & - & (20K, 108K) \\
Terminate on live loss (train, test) & - & (No, Yes) \\

\bottomrule
\end{tabular}
\end{small}
\end{center}
\vskip -0.1in
\end{table}

\begin{table}[h]
\caption{Per-Component Hyperparameters}
\label{table:per-component-hyperparams}
\vskip 0.15in
\begin{center}
\begin{small}
\begin{tabular}{lccccr}
\toprule
Description & Symbol & Tokenizer & World Model & Actor-Critic \\
\midrule
Learning rate & - & 0.0001 & 0.0002 & 0.0001 \\
Batch size & - & 128 & 64 & 128 \\
Gradient Clipping Threshold & - & 10 & 100 & 3 \\
Start after epochs & - & 5 & 25 & 50 \\
Training Steps per epoch & - & 200 & 200 & 100 \\
AdamW Weight Decay & - & 0.01 & 0.05 & 0.01 \\

\bottomrule
\end{tabular}
\end{small}
\end{center}
\vskip -0.1in
\end{table}

\subsubsection{Tokenizer ($\Tokenizer$)}
\label{sec:tokenizer-details}
\paragraph{Tokenizer Architecture}
Our tokenizer is based on the implementation of VQ-GAN \cite{esser2021tamingVQGAN}.
However, we simplified the architectures of the encoder and decoder networks.
A description of the architectures of the encoder and decoder networks can be found in table \ref{table:tokenizer-arch}.

\begin{table}[h]
\caption{The encoder and decoder architectures. ``Conv(a,b,c)" represents a convolutional layer with kernel size $a \times a$, stride of $b$ and padding $c$. A value of $c=\text{Asym.}$ represents an asymmetric padding where a padding of 1 is added only to the right and bottom ends of the image tensor. ``GN" represents a GroupNorm operator with $8$ groups, $\epsilon=1e-6$ and learnable per-channel affine parameters. SiLU is the Sigmoid Linear Unit activation \cite{hendrycks2017bridging, ramachandran2018searching}. ``Interpolate" uses PyTorch's interpolate method with scale factor of 2 and the ``nearest-exact" mode.}
\label{table:tokenizer-arch}
\vskip 0.15in
\begin{center}
\begin{small}
\begin{tabular}{lcr}
\toprule
Module & Output Shape \\
\midrule

Encoder \\
\midrule

Input & $3 \times 64 \times 64$ \\
Conv(3, 1, 1) & $32 \times 64 \times 64$ \\
EncoderBlock1 & $64 \times 32 \times 32$ \\
EncoderBlock2 & $128 \times 16 \times 16$ \\
EncoderBlock3 & $256 \times 8 \times 8$ \\
GN & $256 \times 8 \times 8$ \\
SiLU & $256 \times 8 \times 8$ \\
Conv(3, 1, 1) & $256 \times 8 \times 8$ \\

\midrule

EncoderBlock  \\
\midrule
Input & $c \times h \times w$ \\
GN & $c \times h \times w$ \\
SiLU & $c \times h \times w$ \\
Conv(3, 2, \text{Asym.}) & $2c \times \frac{h}{2} \times \frac{w}{2}$ \\
Conv(3, 1, 1) & $2c \times \frac{h}{2} \times \frac{w}{2}$ \\

\midrule
Decoder \\
\midrule
Input & $256 \times 8 \times 8$ \\
Conv(3, 1, 1) & $256 \times 8 \times 8$ \\
DecoderBlock1 & $128 \times 16 \times 16$ \\
DecoderBlock2 & $64 \times 32 \times 32$ \\
DecoderBlock3 & $32 \times 64 \times 64$ \\
GN & $32 \times 64 \times 64$ \\
SiLU & $32 \times 64 \times 64$ \\
Conv(3, 1, 1) & $3 \times 64 \times 64$ \\

\midrule
DecoderBlock \\
\midrule
Input & $c \times h \times w$ \\
GN & $c \times h \times w$ \\
SiLU & $c \times h \times w$ \\
Interpolate & $c \times 2h \times 2w$ \\
Conv(3, 1, 1) & $\frac{c}{2} \times 2h \times 2w$ \\
Conv(3, 1, 1) & $\frac{c}{2} \times 2h \times 2w$ \\

\bottomrule
\end{tabular}
\end{small}
\end{center}
\vskip -0.1in
\end{table}

\paragraph{Tokenizer Learning}
\label{sec:tokenizer-learining-details}
Following IRIS \cite{micheli2022transformers}, our tokenizer is a VQ-VAE \cite{vanDenOord2017vqvae} based on the implementation of \cite{esser2021tamingVQGAN} (without the discriminator). 
The training objective is given by
\begin{equation}
\label{eq:tokenizer-objective-loss}
    \mathcal{L}(E, D, \tknzrCodebook) = \Vert x - D(z) \Vert_{1} + \Vert 
\text{sg}(E(x)) - \tknzrCodebook(z) \Vert_{2}^{2} + \Vert \text{sg}(\tknzrCodebook(z)) - E(x) \Vert_{2}^{2} + \mathcal{L}_{\text{perceptual}}(x, D(z))
\end{equation}
where $E$ and $D$ are the encoder and decoder models, respectively, and $\text{sg}(\cdot)$ is the stop-gradient operator.
The first term on the right hand side of Equation \ref{eq:tokenizer-objective-loss} above is the reconstruction loss, the second and third terms correspond to the commitment loss, and the last term is the perceptual loss.

\pagebreak

\subsubsection{Retentive World Model ($\WM$)}
\label{sec:wm-details}

The hyperparameters of $\WM$ are presented in Table \ref{table:world-model-arch}.

\paragraph{Implementation Details}
We use the ``Yet-Another-RetNet" RetNet implementation\footnote{https://github.com/fkodom/yet-another-retnet}, as its code is simple and convenient while its performance remain competitive with the official implementation in terms of run time and efficiency.

Originally, the IRIS algorithm provides the world model with a single observation to make forward predictions.
Our implementation considers a context of two frames for making forward predictions.

\begin{table}[h]
\caption{The world model hyper-parameters.}
\label{table:world-model-arch}
\vskip 0.15in
\begin{center}
\begin{small}
\begin{tabular}{lccr}
\toprule
Description & Symbol & Value \\
\midrule

Number of layers & - & 5 \\
Number of Retention heads & - & 4 \\
Embedding dimension & \retnetDmodel & 256 \\
Dropout & - & 0.1 \\
RetNet feed-forward dimension & - & 1024 \\
RetNet LayerNorm epsilon & - & 1e-6 \\
Blocks per chunk & \blocksPerChunk & 3 \\
World model context length & - & 2 \\
\bottomrule
\end{tabular}
\end{small}
\end{center}
\vskip -0.1in
\end{table}

\paragraph{POP Observation-Generation Modes}
The simpler observation generation mode presented in Section \ref{subsec:method-infer} requires two sequential world model calls to generate the next observation. 
The first call consumes the previous observation-action block to compute the recurrent state at the current time step, while the second call uses $\predTokens$ to generate the next observation tokens.
Note that the next observation tokens sampled in the second call have not been processed by the world model at this point.
To incorporate these tokens into the recurrent state, an additional world model call is required.
Here, the cost induced by the first world model call is $\tokensPerObs + 1$, while the second call costs $\tokensPerObs$.

Alternatively, it is also possible to combine these two calls into one by concatenating the previous observation-action block and $\predTokens$.
However, to avoid having $\predTokens$ incorporated in the recurrent state computed by this call, a modified forward call should be used.
Concretely, the resulting recurrent state should only summarize the previous observation-action block, neglecting the suffix $\predTokens$.
In practice, we use the backbone of the POP chunkwise forward mode (Alg. \ref{alg:pop-retnet-fwd-training}) for this computation.
This alternative mode induces only $\horizon$ sequential world model calls, while each call processes $2\tokensPerObs + 1$.
Hence, this alternative reduces the number of sequential calls while maintaining the same total cost.

In practice, the optimal mode to use depends on the configuration, model sizes, and hardware.
In our configuration, we opted for a larger batch size for the imagination phase.
Hence, we found the first (simpler) mode to be slightly more efficient in this case.
However, the second mode could be more efficient in other settings.

\begin{table}[h]
\caption{A comparison between POP and the ``No POP" ablation in terms of their computational costs at training and their number of sequential model forward calls at inference (imagination). For brevity, we only consider observation-prediction related costs, neglecting costs related to the processing of action tokens. POP provides two modes of operation during imagination. Here, we consider the costs for a single input sequence.}
\label{table:pop-vs-no-pop-computational-costs}
\vskip 0.15in
\begin{center}
\begin{small}
\begin{tabular}{lccccr}
\toprule
Algorithm & \makecell[c]{Observation \\ Prediction Cost}  & \makecell[c]{World Model \\ Training Cost}  & \makecell[c]{Imagination \\ Sequential Calls}   &   \makecell[c]{Imagination \\ Cost per Call}   \\
\midrule

POP (default mode) & $2 \tokensPerObs$ & $2 \tokensPerObs \horizon$   &   $2 \horizon$  &  $\tokensPerObs$ \\
POP (alternative mode) & $2 \tokensPerObs$ & $2 \tokensPerObs \horizon$   &   $\horizon$  &  $2 \tokensPerObs$ \\
No POP & $\tokensPerObs$ & $\tokensPerObs \horizon$   &   $\tokensPerObs \horizon$  & $1$ \\

\bottomrule
\end{tabular}
\end{small}
\end{center}
\vskip -0.1in
\end{table}

\subsubsection{Controller  ($\Controller$)}
\label{sec:actor-critic-details}

\paragraph{Actor-Critic Learning}
\label{sec:actor-critic-learning-details}
Our learning algorithm follows IRIS and Dreamer \cite{micheli2022transformers, Hafner2020Dreamerv1, hafner2021dreamerv2}, which uses $\lambda$-returns defined recursively as 
\begin{equation*}
    G_{t} = \begin{cases}
        \hat{r}_t + \gamma (1 - \hat{d}_{t}) \left( (1-\lambda)V^{\pi}(\hat{\tokens}_{t+1}) + \lambda G_{t+1} \right) & t < H \\
        V^{\pi}(\hat{\tokens}_{H}) & t=H
    \end{cases}
\end{equation*}
where $V^{\pi}$ is the value network learned by the critic, and $(\hat{\tokens}_0, \action_0 , \hat{\reward}_{0}, \hat{\doneSgnl}_{0}, \ldots, \hat{\tokens}_{H-1}, \action_{H-1} , \hat{\reward}_{H-1}, \hat{\doneSgnl}_{H-1}, \hat{\tokens}_{H})$ is a trajectory obtained through world model imagination.

To optimize $V^{\pi}$, the following loss is minimized:
\begin{equation*}
    \mathcal{L}_{V^{\pi}} = \mathbb{E}_{\pi} [ \sum_{t=0}^{H-1} V^{\pi}(\hat{\tokens}_{t}) - \text{sg}(G_t)^2 ]
\end{equation*}

The policy optimization follows a simple REINFORCE \cite{sutton2018reinforcement} objective, with $V^{\pi}$ used as a baseline for variance reduction.
The objective is given by
\begin{equation*}
    \mathcal{L}_{\pi} = -\mathbb{E}_{\pi} [\sum_{t=0}^{H-1} \log(\pi(a_t \vert \hat{\tokens}_{1}, \action_1, \ldots, \hat{\tokens}_{t-1}, \action_{t-1}, \hat{\tokens}_{t} )) \text{sg}(G_t - V^{\pi}(\hat{\tokens}_{t})) + \alpha \mathcal{H}(\pi(a_t \vert \hat{\tokens}_{1}, \action_1, \ldots, \hat{\tokens}_{t-1}, \action_{t-1}, \hat{\tokens}_{t} )) ]
\end{equation*}
where $\alpha$
The values of the hyperparameters used in our experiments are detailed in Table \ref{table:actor-critic-hparams}

\begin{table}[h]
\caption{Actor-Critic Hyperparameters.}
\label{table:actor-critic-hparams}
\vskip 0.15in
\begin{center}
\begin{small}
\begin{tabular}{lccr}
\toprule
Description & Symbol & Value \\
\midrule

Discount factor & $\gamma$ & 0.995 \\
$\lambda$-return & $\lambda$ & 0.95 \\
Entropy loss weight & $\alpha$ & 0.001 \\

\bottomrule
\end{tabular}
\end{small}
\end{center}
\vskip -0.1in
\end{table}

\begin{table}[h]
\caption{The actor-critic observation representation architecture.}
\label{table:actor-critic-obs-repr-arch}
\vskip 0.15in
\begin{center}
\begin{small}
\begin{tabular}{lcr}
\toprule
Module & Output Shape \\
\midrule

Input & $256 \times 8 \times 8$ \\
Conv(3, 1, 1) & $128 \times 8 \times 8$ \\
SiLU & $128 \times 8 \times 8$ \\
Conv(3, 1, 1) & $64 \times 8 \times 8$ \\
SiLU & $64 \times 8 \times 8$ \\
Flatten & 4096 \\
Linear & 512 \\
SiLU & 512 \\

\bottomrule
\end{tabular}
\end{small}
\end{center}
\vskip -0.1in
\end{table}

\paragraph{Agent Architecture}
\label{sec:actor-critic-arch}
The architecture of the agent module comprises of a shared backbone and two linear maps for the actor and critic heads, respectively.
The shared backbone first maps the input to a latent representation which takes the form of a 512-dimensional vector.
For action token inputs, a learned embedding table is used to map the token to its latent representation.
For observation inputs, the $\tokensPerObs$ tokens are first mapped to their corresponding code vectors learned by the tokenizer and reshaped according to their original spatial order.
Then, the resulting tensor is processed by a convolutional neural network followed by a fully connected network.
The architecture details of these networks are presented in Table \ref{table:actor-critic-obs-repr-arch}.
Lastly, a long-short term memory (LSTM) \cite{hochreiter1997lstm} network of dimension $512$ maps the processed input vector to an history-dependant latent vector, which serves as the output of the shared backbone.

\paragraph{Action Dependant Actor Critic}
In the IRIS algorithm, the actor and critic networks share an LSTM \cite{hochreiter1997lstm} backbone and model $\policy(\action_{t} \vert \obs_{\leq t}), \valueFn( \obs_{\leq t} )$.
Notice that the output of the policy models the \emph{distribution} of actions at step $t$.
Importantly, the model has no information about the sampled actions.
In REM, the input of $\Controller$ contains the sampled actions, i.e., our algorithm models $\policy(\action_{t} \vert \hat{\tokens}_{1}, \action_1, \ldots, \hat{\tokens}_{t-1}, \action_{t-1}, \hat{\tokens}_{t}), \valueFn(\hat{\tokens}_{1}, \action_1, \ldots, \hat{\tokens}_{t-1}, \action_{t-1}, \hat{\tokens}_{t})$.

\clearpage

\subsection{REM Algorithm}
\label{sec:rem-algorithm-pseudo-code}
Here, we present a pseudo-code of REM.
The high-level loop is presented in Algorithm \ref{alg:overview-alg}, while the pseudo-codes of the training of each component are presented in algorithms \ref{alg:overview-alg-collect-exp}-\ref{alg:overview-alg-train-c}.

\begin{algorithm}[h]
   \caption{REM Training Overview}
   \label{alg:overview-alg}
\begin{algorithmic}
   \STATE {\bfseries Input:} 
   \REPEAT
   \STATE $\texttt{collect\_experience()}$ (Alg. \ref{alg:overview-alg-collect-exp})
   \STATE $\texttt{train\_V()}$ (Alg. \ref{alg:overview-alg-train-v})
   \STATE $\texttt{train\_M()}$ (Alg. \ref{alg:overview-alg-train-m})
   \STATE $\texttt{train\_C()}$ (Alg. \ref{alg:overview-alg-train-c})
   \UNTIL{stopping criterion is met}
\end{algorithmic}
\end{algorithm}

\begin{algorithm}[h]
   \caption{$\texttt{collect\_experience}$}
   \label{alg:overview-alg-collect-exp}
\begin{algorithmic}
   \STATE {\bfseries Input:} 
   \STATE $\obs_1 \leftarrow \texttt{env.reset()}$
   \FOR{$t=1$ {\bfseries to} $T$}
   
   \STATE $\tokens_{t} \leftarrow \Tokenizer_{\text{Enc}}(\obs_{t})$
   \STATE $\action_{t} \sim \policy(a_{t} \vert \tokens_{1}, \action_{1}, \ldots, \tokens_{t-1}, \action_{t-1}, \tokens_{t})$
    \STATE $\obs_{t+1}, \reward_{t}, \doneSgnl_{t} \leftarrow \texttt{env.step}(\action_t)$
    \IF{$\doneSgnl_{t} = 1$}
    \STATE $\obs_{t+1} \leftarrow \texttt{env.reset()}$
    \ENDIF

   \ENDFOR

   \STATE $\texttt{replay\_buffer.store}( \{ \obs_{t} , \action_{t} , \reward_{t} , \doneSgnl_{t} \}_{t=1}^{T} )$
   
\end{algorithmic}
\end{algorithm}

\begin{algorithm}[h]
   \caption{$\texttt{train\_V}$}
   \label{alg:overview-alg-train-v}
\begin{algorithmic}
   \STATE $\obs  \leftarrow \texttt{replay\_buffer.sample\_obs()}$
   \STATE $\tokens \leftarrow \Tokenizer_\text{Enc}(\obs)$
   \STATE $\hat{\obs} \leftarrow \Tokenizer_{\text{Dec}}(\tokens)$
   \STATE Compute loss (Eqn. \ref{eq:tokenizer-objective-loss})
   \STATE Update $\Tokenizer$

\end{algorithmic}
\end{algorithm}

\begin{algorithm}[t]
   \caption{$\texttt{train\_M}$}
   \label{alg:overview-alg-train-m}
\begin{algorithmic}
   \STATE $\{ \obs_{t} , \action_{t} , \reward_{t} , \doneSgnl_{t} \}_{t=1}^{\horizon} \leftarrow \texttt{replay\_buffer.sample()}$
   
   \FOR{$t=1$ {\bfseries to} $\horizon$}

   \STATE $\tokens_{t}, \tknzrLatent_{t} \leftarrow \Tokenizer_{\text{Enc}}(\obs_{t})$
   \STATE $\actionEmb_{t} \leftarrow \WM\texttt{.embed\_action}(\action_{t})$

   \ENDFOR

   \STATE $\retnetInput \leftarrow (\tknzrLatent_{1}, \actionEmb_{1}, \ldots, \tknzrLatent_{\horizon}, \actionEmb_{\horizon}) $

   \STATE $\retnetState_{0}^{1}, \ldots, \retnetState_{0}^{L} \leftarrow 0, \ldots, 0$

    \FOR{$i=1$ {\bfseries to}  $\lceil \frac{\horizon}{\blocksPerChunk} \rceil$}
    
    \STATE $\retnetY_{[i]}, \{\retnetState_{i}^{l}\}_{l=1}^{L} \leftarrow \texttt{POP\_forward}(\retnetInput_{[i]}, \{\retnetState_{[i-1]}^{l}\}_{l=1}^{L})$ (Alg. \ref{alg:pop-retnet-fwd-training})
    
    \ENDFOR
    
   \STATE $(\hat{\tokens}_{1}, \ldots, \hat{\tokens}_{\horizon}) \leftarrow \WM\texttt{.obs\_pred\_head}(\retnetY\texttt{[:, :-1]})$
   
   \STATE $(\hat{\reward}_{1}, \hat{\doneSgnl}_{1}, \ldots, \hat{\reward}_{\horizon}, \hat{\doneSgnl}_{\horizon}) \leftarrow \WM\texttt{.reward\_done\_head}(\retnetY\texttt{[:, -1]})$

   \STATE Compute Losses and update $\WM$
   
\end{algorithmic}
\end{algorithm}

\begin{algorithm}[t]
   \caption{$\texttt{train\_C}$}
   \label{alg:overview-alg-train-c}
\begin{algorithmic}
   \STATE $\{ \obs_{t} , \action_{t} , \reward_{t} , \doneSgnl_{t} \}_{t=1}^{\horizon} \leftarrow \texttt{replay\_buffer.sample()}$

    \FOR{$t=1$ {\bfseries to} $\horizon$}

   \STATE $\tokens_{t}, \tknzrLatent_{t} \leftarrow \Tokenizer_{\text{Enc}}(\obs_{t})$

   \ENDFOR

   \STATE $\retnetState \leftarrow 0$
   \STATE $\blocksPerChunk \leftarrow 1$

    

   \STATE Initialize context $\tau \leftarrow (\tokens_{1}, \action_{1}, \dots, \tokens_{\horizon})$

   
    \FOR{$t'=\horizon+1$ {\bfseries to} $2\horizon$}
    \STATE $\action_{t'} \sim \pi(\action_{t'} \vert \tokens_{1}, \action_{1}, \ldots, \tokens_{t'})$
    \STATE $V_{t'} \leftarrow V(\tokens_{1}, \action_{1}, \ldots, \tokens_{t'})$


    \STATE $\retnetY, \retnetState \leftarrow \WM\texttt{.retnet\_chunkwise\_forward}((\tau, \action_{t'}), \retnetState, t')$

    \STATE $\reward_{t'}, \doneSgnl_{t'} \sim \WM\texttt{.reward\_done\_head}(\retnetY\texttt{[-1]})$

    \STATE $\retnetY, \text{\_} \leftarrow \WM\texttt{.retnet\_chunkwise\_forward}(\predTokens, \retnetState, t'+1)$
    \STATE $\tokens_{t'+1} \sim \WM\texttt{.obs\_pred\_head}(\retnetY\texttt{[:-1]})$

    \STATE $\tau \leftarrow \tokens_{t'+1}$
    

    \ENDFOR

    \STATE Update $\pi, V$ (detailed in Section \ref{sec:actor-critic-learning-details})

\end{algorithmic}
\end{algorithm}

\clearpage

\subsection{Retentive Networks}
\label{sec:retentive-nets-details}

In this section, we give detailed information regarding the RetNet architecture for the completeness of this paper.
For convenience reasons, we defer to the notations of \cite{sun2023retentive}, rather than the notation presented in 

\cite{sun2023retentive} is a recent alternative to Transformers \cite{Vaswani2017Attention}. It is highly parallelizable, has lower cost inference than Transformers, and is empirically claimed to perform competitively on language modelling tasks.
The RetNet model is a stack of $\retnetNumLayers$ identical layers.
Here, we denote the output of the $l$-th layer by $\retnetLayerOutput{l}$.
Given an embedded input sequence $\retnetInput = \retnetLayerOutput{0} \in \mathbb{R}^{\retnetSeqLen \times \retnetDmodel}$ of $\retnetSeqLen$ $\retnetDmodel$-dimensional vectors, each RetNet layer can be described as
\begin{gather}
    \retnetLayerMSROut^{l} = \retnetLayerMSR(\retnetLayerLN(\retnetLayerOutput{l})) + \retnetLayerOutput{l} \\
    \retnetLayerOutput{l+1} = \retnetLayerFFN(\retnetLayerLN(\retnetLayerMSROut^{l})) + \retnetLayerMSROut^{l}
\end{gather}
where  $\retnetLayerLN(\cdot)$ is layer-norm \cite{ba2016layer}, $\retnetLayerFFN(\retnetInput) = \retnetGeluActivation(\retnetInput \textbf{W}_1 ) \textbf{W}_2$ is a feed-forward network (FFN), and $\retnetLayerMSR(\cdot)$ is a multi-scale retention (MSR) module with multiple Retention heads.
The output of the RetNet model is given by $\retnetOp{\retnetLayerOutput{0}} = \retnetLayerOutput{\retnetNumLayers}$.

As presented in the main text, the chunkwise equations are 
\begin{gather*}
    \retnetState_{[i]} = (\retnetK_{[i]} \odot \boldsymbol{\zeta})^{\Tr} \retnetV_{[i]} + \retnetEta^{\retnetChunkSize} \retnetState_{[i-1]} \\
         \retnetY_{[i]} = \left( \retnetQ_{[i]} \retnetK^{\Tr}_{[i]} \odot \retnetD \right) \retnetV_{[i]}   +    (\retnetQ_{[i]} \retnetState_{[i-1]}) \odot \boldsymbol{\xi}
\end{gather*}
where $\retnetQ = \left( \retnetInput \retnetW_{Q} \right) \odot \retnetPos$, $\retnetK =  \left( \retnetInput \retnetW_{K} \right) \odot \retnetPosC$, $\retnetV = \retnetInput \retnetW_{V}$, and $\boldsymbol{\xi}\in\mathbb{R}^{B\times d}$ is a matrix with $\boldsymbol{\xi}_{ij} = \retnetEta^{i+1}$.
Here, $\retnetPos_n =  e^{\iu n \retnetPosTheta }$, $\retnetD_{n,m} = \begin{cases}
    \retnetEta^{n-m} & n \geq m \\
    0 & n < m
\end{cases}$,  and $\retnetPosTheta, \retnetEta \in \mathbb{R}^{\retnetDmodel}$, 

where $\retnetEta$ is an exponential decay factor, and the matrices $\retnetPos,\retnetPosC\in\mathbb{C}^{m\times d}$ are for relative position encoding, and $\retnetD\in\mathbb{R}^{\retnetChunkSize \times \retnetChunkSize}$ combines an auto-regressive mask with the temporal decay factor $\retnetEta$.

In each RetNet layer, $h = \frac{d}{d_{\text{head}}}$ heads are used, where $d_{\text{head}}$ is the dimension of each head.
Head Retention head uses different parameters $W_{K}, W_{Q}, W_{V}$.
Additionally, Retention head uses a different value of $\retnetEta$.
Among different RetNet layers, the values of $\retnetEta$ are fixed.
Each layer is defined as follows:
\begin{gather*}
    \retnetEta = 1- 2^{-5-\text{arange}(0, h)} \in \mathbb{R}^{h} \\
    \text{head}_i = \text{Retention}(X, \retnetEta_i) \\
    Y = \text{GroupNorm}_{h}(\text{Concat}(\text{head}_{1}, \ldots, \text{head}_{h})) \\
    \text{MSR}(X) = (\text{swish}(XW_{G}) \odot Y) W_O
\end{gather*}
where $W_G, W_O \in \mathbb{R}^{d\times d}$ are learnable parameters.

\clearpage

\subsection{Additional Results}
\label{sec:additional-results}

\begin{figure}[h]
    \centering
    \includegraphics[width=0.5\linewidth]{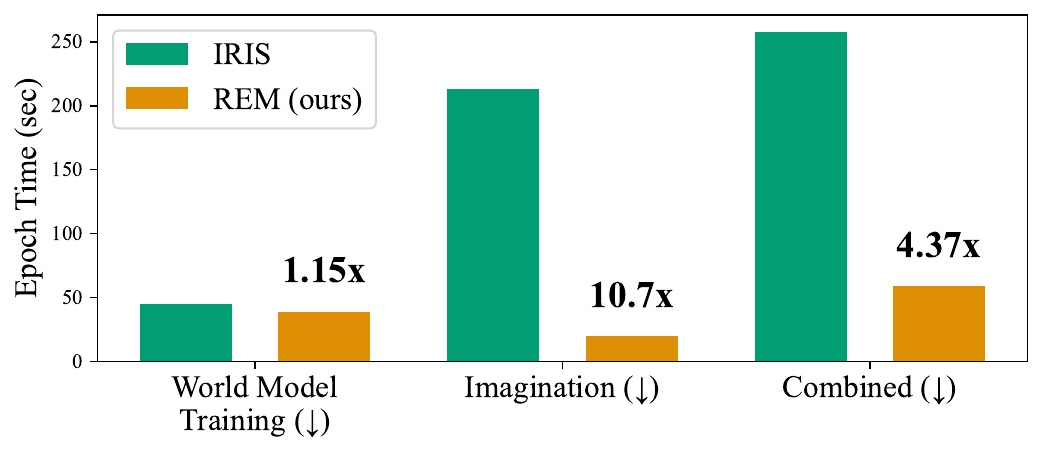}
    \caption{A comparison between the run times of REM and an improved version of IRIS that uses REM's configurations (detailed in \ref{sec:models-and-hyperparameters}) during the world model training and imagination phases (actor-critic training). }
    \label{fig:time-comparison-fair}
\end{figure}

In addition to comparing the run times of REM and IRIS, we also conducted a comparison to an improved version of IRIS that uses REM's configurations.
These results are presented in Figure \ref{fig:time-comparison-fair}.
These results clearly show the effectiveness of our novel POP mechanism.

\begin{figure}[h]
    \centering
    \includegraphics[width=0.35\linewidth]{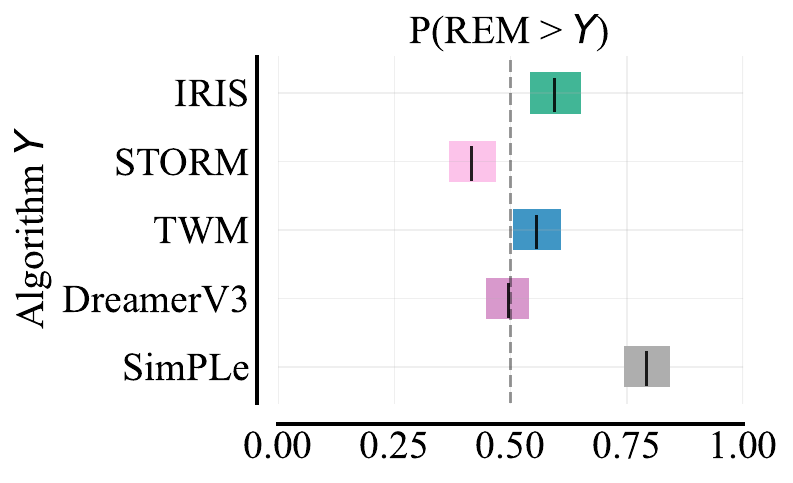}
    \caption{The probability of improvement \cite{Agarwal2021rliable} shows the probability of REM outperforming each baseline on a randomly selected game from the 26 games of Atari 100K with 95\% stratified bootstrap confidence intervals.}
    \label{fig:main-exp-proba-of-improvement}
\end{figure}

The probability of improvement results from our Atari 100K benchmark experiment are presented in Figure \ref{fig:main-exp-proba-of-improvement}.
Importantly, REM outperforms previous token-based methods, namely, IRIS, while competitive with all baselines except STORM on this metric.
We highlight that our main contributions address the computational bottleneck of token-based methods, and thus we focus on comparing REM to these approaches. 

\begin{figure*}[h]
    \centering
    \includegraphics[width=\linewidth]{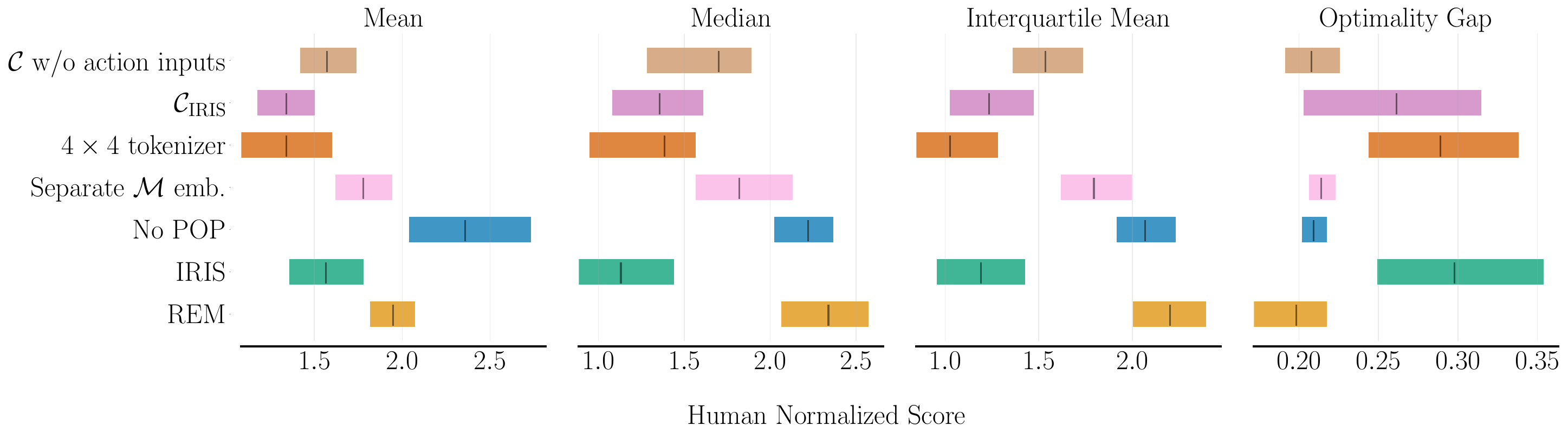}
    \caption{Aggregated metrics with 95\% stratified bootstrap confidence intervals of the mean, median, and inter-quantile mean (IQM) human-normalized scores and optimality gap \cite{Agarwal2021rliable} for each ablation on a subset of 8 games from the Atari 100K benchmark. The results are based on 5 random seeds.}
    \label{fig:ablations-aggregates-full}
\end{figure*}

\begin{figure}[h]
    \centering
    \includegraphics[width=0.45\linewidth]{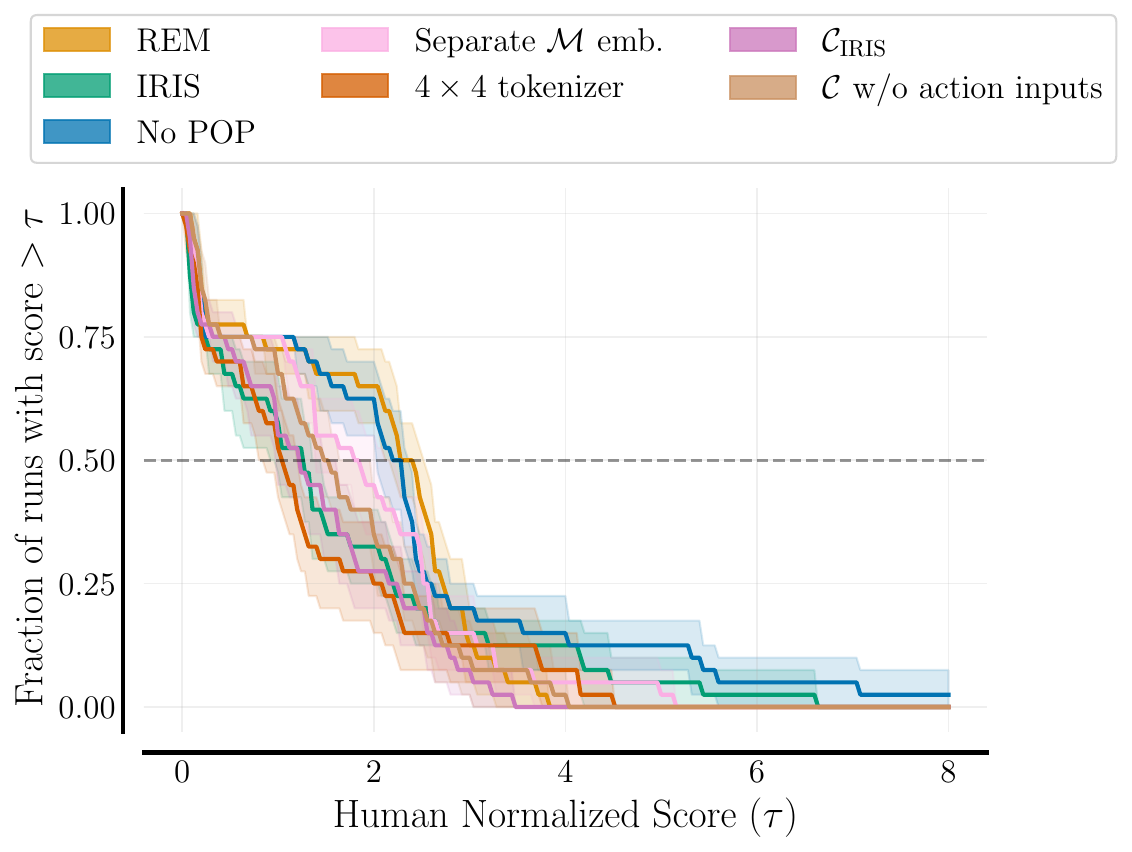}
    \caption{The Performance profiles of the ablations. For every human-normalized score value (x axis), each algorithm's curve shows the fraction of its runs with score grater than the given score value. The shaded area indicates pointwise 95\% confidence bands based on percentile bootstrap with stratified sampling  \cite{Agarwal2021rliable}. The results of each algorithm and each game from the subset of 8 Atari games used in our ablations are based on 5 random seeds}
    \label{fig:ablations-performance-profile}
\end{figure}

The complete set of ablation results are presented in Figure \ref{fig:ablations-aggregates-full} and Table \ref{table:ablations-full-results}.
The performance profiles for the ablations are presented in Figure \ref{fig:ablations-performance-profile}.

\begin{table*}[h]
\caption{Mean agent returns on a subset of 8 games from the Atari 100k benchmark followed by averaged human-normalized performance metrics. Each game score is computed as the average of 5 runs with different seeds, where the score of each run is computed as the average over 100 episodes sampled at the end of training. The best score on each game is indicated with bold face. }
\label{table:ablations-full-results}
\vskip 0.15in
\begin{center}
\begin{small}
\begin{tabular}{lcc ccccc cr}
\toprule

Game                 &  Random    &  Human     &  REM         & IRIS        &  No POP              &  \makecell[c]{Separate \\ {\ttfamily $\WM$} emb.}   &  \makecell[c]{$4 \times 4$ \\ tokenizer}   &  \makecell[c]{{ $\Controller_{\text{IRIS}}$}}   &  \makecell[c]{$\Controller$ w/o \\ action inputs}  \\
\midrule
Assault              &  222.4     &  742.0     &  \textbf{1764.2}    &  1524.4            &  1472.2             &  1269.2                               &  1288.9                   &  1221.5                               &  1498.5                                  \\
Asterix              &  210.0     &  8503.3    &  1637.5             &  853.6             &  1603.4             &  1185.9                               &  909.6                    &  1376.4                               &  \textbf{1656.3}                         \\
ChopperCommand       &  811.0     &  7387.8    &  \textbf{2561.2}    &  1565.0            &  1848.0             &  1928.3                               &  1958.9                   &  2517.6                               &  2302.7                                  \\
CrazyClimber         &  10780.5   &  35829.4   &  \textbf{76547.6}   &  59324.2           &  62964.8            &  74791.3                              &  57814.7                  &  30952.7                              &  42441.2                                 \\
DemonAttack          &  152.1     &  1971.0    &  5738.6             &  2034.4            &  \textbf{12316.0}   &  4389.9                               &  3863.3                   &  5159.0                               &  5827.0                                  \\
Gopher               &  257.6     &  2412.5    &  \textbf{5452.4}    &  2236.1            &  5338.4             &  3764.2                               &  2174.9                   &  2891.2                               &  4365.3                                  \\
Krull                &  1598.0    &  2665.5    &  4017.7             &  \textbf{6616.4}   &  5138.6             &  5779.9                               &  4612.8                   &  3866.2                               &  3659.6                                  \\
RoadRunner           &  11.5      &  7845.0    &  \textbf{14060.2}   &  9614.6            &  13161.6            &  11723.5                              &  6161.7                   &  11692.9                              &  11692.9                                 \\
\midrule
\#Superhuman (↑)     &  0         &  N/A       &  \textbf{6}         &  5                 &  6                  &  6                                    &  4                        &  5                                    &  6                                       \\
Mean (↑)             &  0.000     &  1.000     &  1.947              &  1.564             &  \textbf{2.357}     &  1.778                                &  1.341                    &  1.340                                &  1.571                                   \\
Median (↑)           &  0.000     &  1.000     &  \textbf{2.339}     &  1.130             &  2.221              &  1.821                                &  1.384                    &  1.357                                &  1.699                                   \\
IQM (↑)              &  0.000     &  1.000     &  \textbf{2.201}     &  1.191             &  2.068              &  1.794                                &  1.026                    &  1.234                                &  1.535                                   \\
Optimality Gap (↓)   &  1.000     &  0.000     &  \textbf{0.198}     &  0.298             &  0.209              &  0.214                                &  0.289                    &  0.261                                &  0.208                                   \\

\bottomrule
\end{tabular}
\end{small}
\end{center}
\vskip -0.1in
\end{table*}

\begin{figure}[h]
    \centering
    \includegraphics[width=0.45\linewidth]{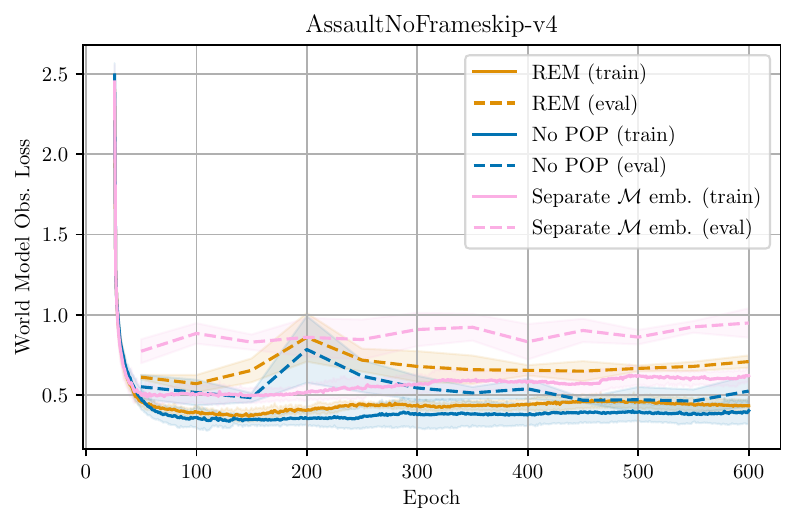}
    \includegraphics[width=0.45\linewidth]{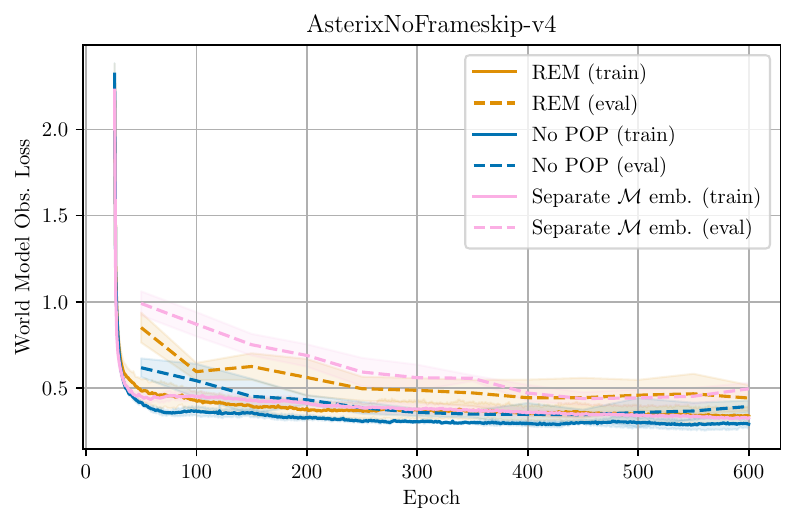}
    \includegraphics[width=0.45\linewidth]{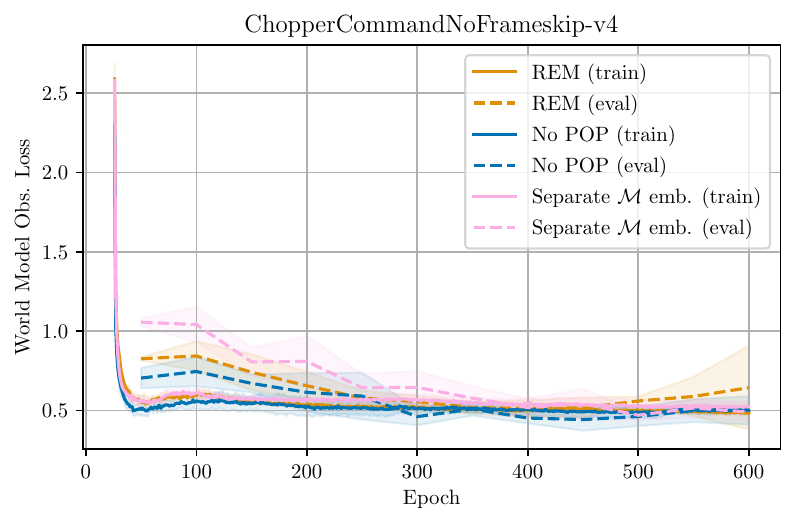}
    \includegraphics[width=0.45\linewidth]{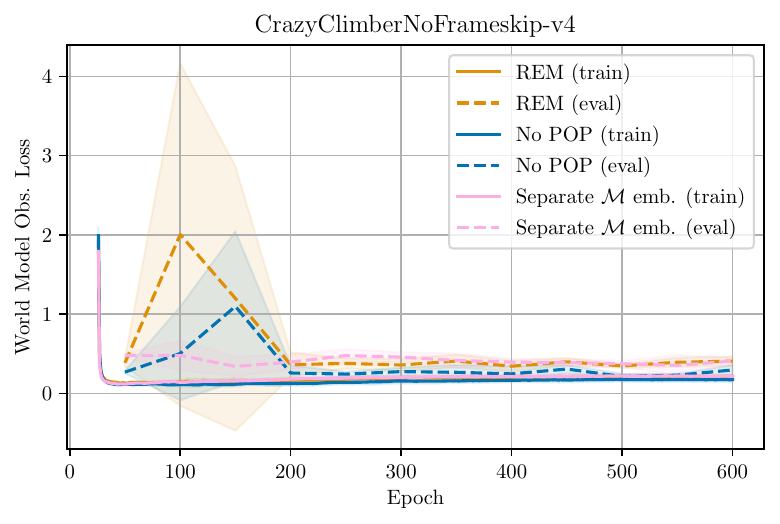}
    \includegraphics[width=0.45\linewidth]{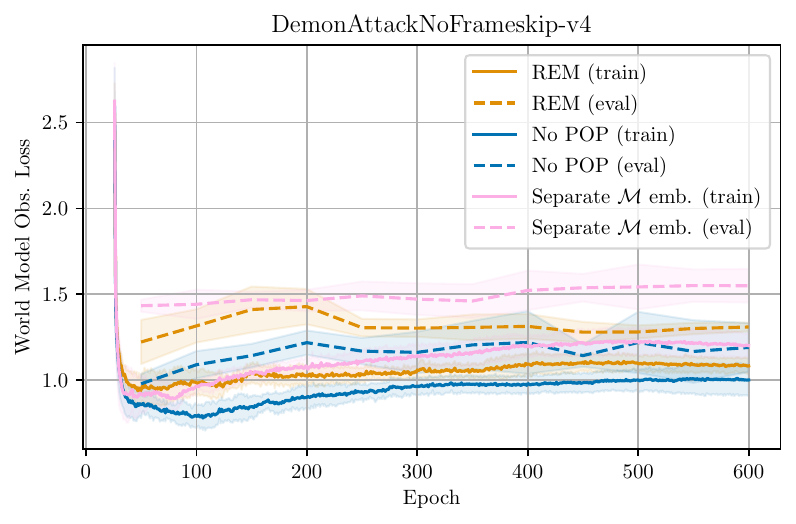}
    \includegraphics[width=0.45\linewidth]{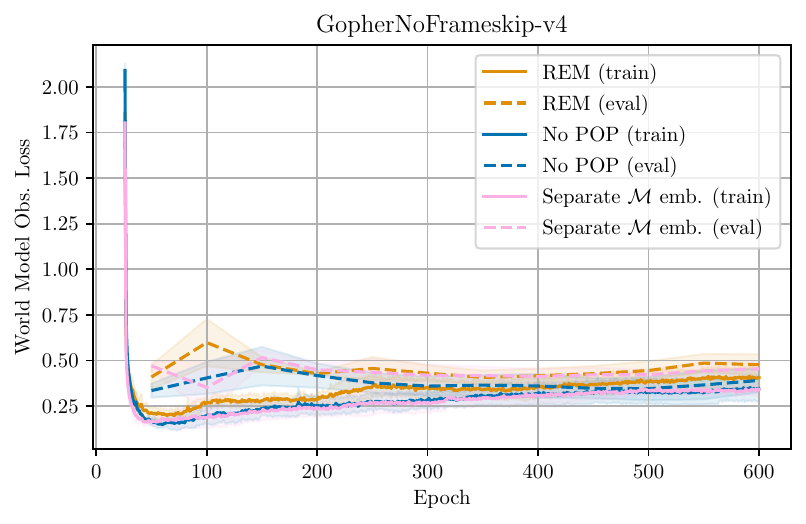}
    \includegraphics[width=0.45\linewidth]{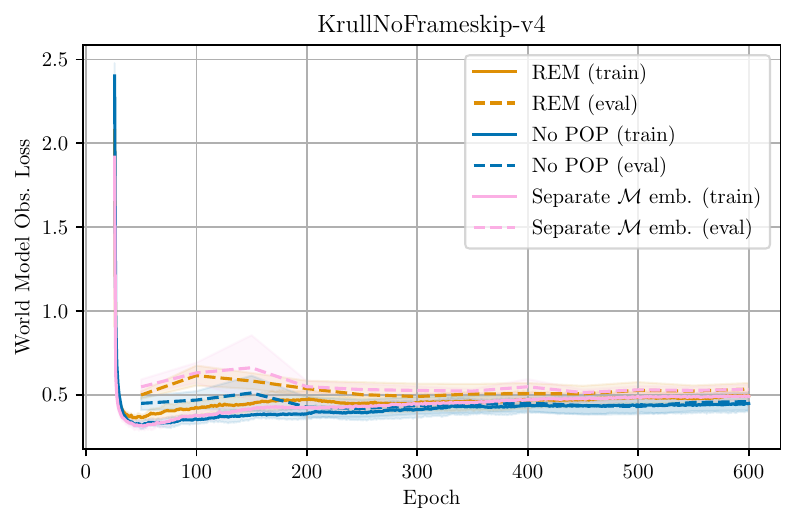}
    \includegraphics[width=0.45\linewidth]{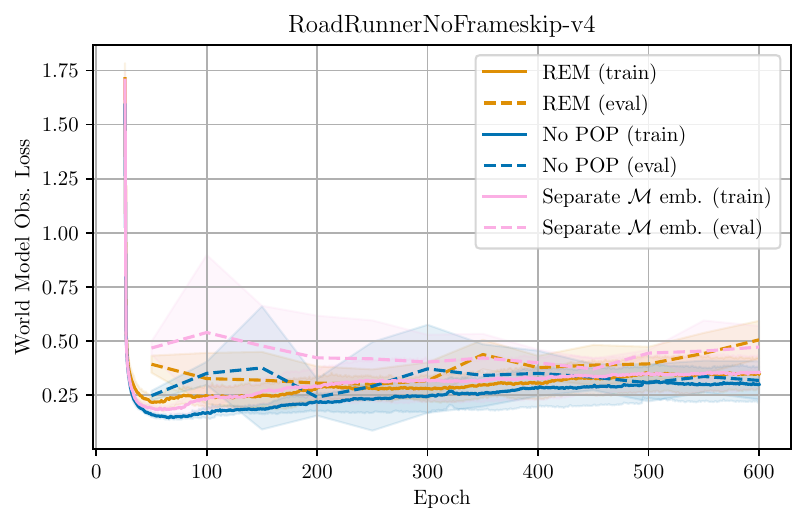}
    \caption{Comparison between the world model observation prediction loss of REM and two of its ablations for each of the 8 games considered in our ablations. For each algorithm, the mean and standard deviation of the training and evaluation losses are shown. The observation prediction loss is computed as the average observation token cross-entropy loss. Note that the evaluation frequency is 50 epochs.}
    \label{fig:ablations-wm-obs-loss-all}
\end{figure}

\paragraph{Ablations World Model Observation Prediction Losses}
To investigate the contribution of each ablation to the quality of world model observation predictions, we measured the corresponding loss values during training and during test episodes with a frequency of 50 epochs.
The results are presented in Figure \ref{fig:ablations-wm-obs-loss-all}, including results for each of the 8 games used in our ablation studies.

\clearpage

\subsection{Setup in Freeway}
For a fair comparison, we followed the actor-critic configurations of IRIS \cite{micheli2022transformers} for Freeway.
Specifically, the sampling temperature of the agent is modified from 1 to 0.01, a heuristic that guides the agent towards non-zero reward trajectories.
We highlight that different methods use other mechanisms such as epsilon-greedy schedules and ``argmax" action selection policies to overcome this exploration challenge \cite{micheli2022transformers}.

\end{document}